\documentclass[sigconf]{acmart}
\usepackage{booktabs} % For formal tables

\usepackage{bm}
\usepackage{mathtools}
\usepackage{float}
\usepackage[ruled,vlined,linesnumbered]{algorithm2e}
\usepackage{algpseudocode}
\usepackage{leftidx}
\usepackage{url}
\usepackage{tabularx, multirow, multicol}
\usepackage{color}
\usepackage{soul}
\usepackage{caption}
\usepackage{subcaption}
\usepackage{float}
\usepackage{balance}
\usepackage{setspace}
\usepackage{bbm}
\usepackage{pifont}

\setcopyright{rightsretained}
\newcolumntype{P}{>{\centering\arraybackslash}m{0.095\linewidth}}
\newcolumntype{Q}{>{\centering\arraybackslash}m{0.1\linewidth}}
\newcolumntype{R}{>{\centering\arraybackslash}m{0.16\linewidth}}
\newcolumntype{S}{>{\centering\arraybackslash}m{0.13\linewidth}}
\newcolumntype{T}{>{\centering\arraybackslash}m{0.2\linewidth}}
\newcolumntype{V}{>{\raggedright\arraybackslash}m{0.42\linewidth}}
\newcolumntype{U}{>{\centering\arraybackslash}m{0.12\linewidth}}

\copyrightyear{2022}
\acmYear{2022}
\setcopyright{acmcopyright}\acmConference[WWW '22]{Proceedings of the ACM Web
Conference 2022}{April 25--29, 2022}{Virtual Event, Lyon, France}
\acmBooktitle{Proceedings of the ACM Web Conference 2022 (WWW '22), April
25--29, 2022, Virtual Event, Lyon, France}
\acmPrice{15.00}
\acmDOI{10.1145/3485447.3512002}
\acmISBN{978-1-4503-9096-5/22/04}

\author{Dongha Lee$^1$, Jiaming Shen$^2$, SeongKu Kang$^3$, Susik Yoon$^1$, Jiawei Han$^1$, Hwanjo Yu$^3$}
\affiliation{%
  \institution{$^1$University of Illinois at Urbana-Champaign, IL, USA \hspace{3pt} $^2$Google Research, NY, USA}
  \institution{$^3$Pohang University of Science and Technology, Republic of Korea}
  \{donghal, susik, hanj\}@illinois.edu, jmshen@google.com, \{seongku, hwanjoyu\}@postech.ac.kr
}

\newenvironment{talign*}
 {\csname align*\endcsname}
 {\endalign}

\newcommand{\segphrase}{SegPhrase\xspace}
\newcommand{\autophrase}{AutoPhrase\xspace}
\newcommand{\caseolap}{CaseOLAP\xspace}

\newcommand{\taxoa}{\mathcal{H}_a\xspace}
\newcommand{\taxob}{\mathcal{H}_b\xspace}
\newcommand{\taxoc}{\mathcal{H}_c\xspace}

\newcommand{\hlda}{hLDA\xspace}
\newcommand{\weshclass}{WeSHClass\xspace}

\newcommand{\corel}{CoRel\xspace}

\newcommand{\taxogen}{TaxoGen\xspace}
\newcommand{\nettaxo}{NetTaxo\xspace}
\newcommand{\proposed}{TaxoCom\xspace}
\newcommand{\josh}{JoSH\xspace}

\newcommand{\nyt}{NYT\xspace}
\newcommand{\arxiv}{arXiv\xspace}

\DeclareMathOperator*{\argmin}{argmin}
\DeclareMathOperator*{\argmax}{argmax}
\DeclareMathOperator*{\Motimes}{\text{\raisebox{0.25ex}{\scalebox{0.8}{$\bigotimes$}}}}

\theoremstyle{definition}
\newtheorem{definition}{Definition}

\newcommand{\svec}[1]{\bm{s}_{#1}}

\newcommand{\tvec}[1]{\bm{t}_{#1}}
\newcommand{\vvec}[1]{\tilde{\bm{t}}_{#1}}

\newcommand{\termset}{\mathcal{T}}
\newcommand{\docuset}{\mathcal{D}}

\newcommand{\taxo}{\mathcal{H}}

\newcommand{\topicnterms}[1]{\mathcal{T}^n_{#1}}
\newcommand{\topickterms}[1]{\mathcal{T}^k_{#1}}
\newcommand{\topicterms}[1]{\mathcal{T}_{#1}}
\newcommand{\topicdocs}[1]{\mathcal{D}_{#1}}

\newcommand{\topicchilds}[1]{\mathcal{S}_{#1}}
\newcommand{\novelchilds}[1]{\mathcal{N}_{#1}}

\newcommand{\smallsection}[1]{{\vspace{0.03in} \noindent \bf {#1.\hspace{5pt}}}}

\begin{document}
\title{\proposed: Topic Taxonomy Completion with Hierarchical Discovery of Novel Topic Clusters}

\begin{abstract}
Topic taxonomies, which represent the latent topic (or category) structure of document collections, provide valuable knowledge of contents in many applications such as web search and information filtering.
Recently, several unsupervised methods have been developed to automatically construct the topic taxonomy from a text corpus, but it is challenging to generate the desired taxonomy without any prior knowledge.
In this paper, we study how to leverage the partial (or incomplete) information about the topic structure as guidance to find out the complete topic taxonomy.
We propose a novel framework for topic taxonomy completion, named \proposed, which recursively expands the topic taxonomy by discovering novel sub-topic clusters of terms and documents.
To effectively identify novel topics within a hierarchical topic structure, \proposed devises its embedding and clustering techniques to be closely-linked with each other:
(i) \textit{locally discriminative embedding} optimizes the text embedding space to be discriminative among known (i.e., given) sub-topics, and (ii) \textit{novelty adaptive clustering} assigns terms into either one of the known sub-topics or novel sub-topics.
Our comprehensive experiments on two real-world datasets demonstrate that \proposed not only generates the high-quality topic taxonomy in terms of term coherency and topic coverage but also outperforms all other baselines for a downstream task.
\end{abstract}



\maketitle

\section{Introduction}
\label{sec:intro}
\begin{figure}[t]
    \centering
    \includegraphics[width=\linewidth]{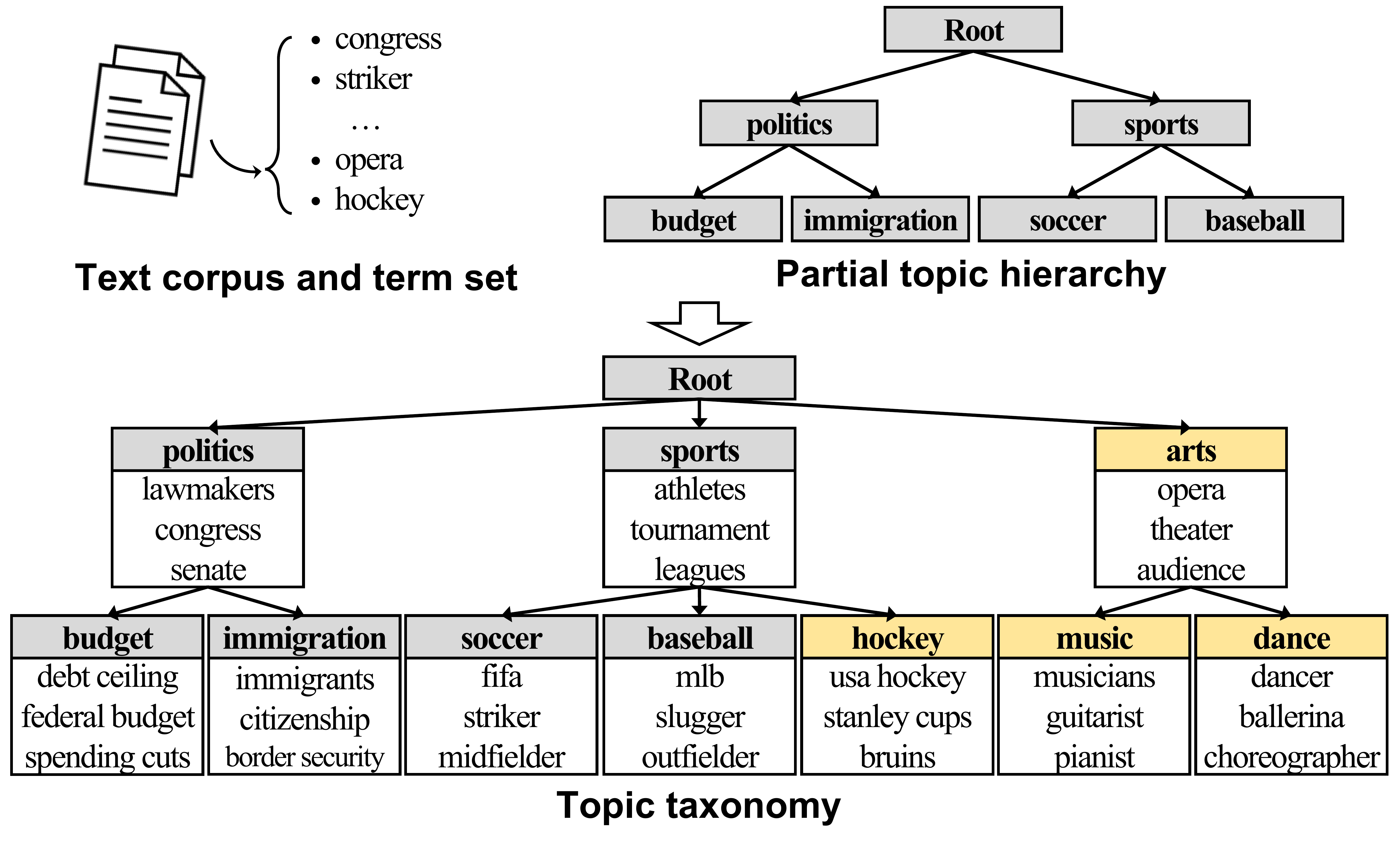}
    \caption{An example of topic taxonomy completion. The known (i.e., given) sub-topics and novel sub-topics are colored in grey and yellow, respectively.}
    \label{fig:problem}
\end{figure}

% 1. Hierarcihcal topic discovery => Unsupervised methods
Finding the latent topic structure of an input text corpus, also known as hierarchical topic discovery~\cite{zhang2018taxogen, shang2020nettaxo, downey2015efficient, wang2013phrase, liu2012automatic}, has been one of the most important problems for information extraction and semantic analysis of text data.
Recently, several studies have focused on topic taxonomy construction~\cite{zhang2018taxogen,shang2020nettaxo}, which aims to generate a tree-structured taxonomy whose node corresponds to a conceptual topic;
each node of the topic taxonomy is defined as a cluster of semantically coherent terms representing a single topic.
Compared to a conventional entity (or term-level) taxonomy, this cluster-level taxonomy is more appropriate for representing the topic hierarchy of the target corpus with high coverage and low redundancy.
To identify hierarchical topic clusters of terms, they mainly performed clustering on a low-dimensional text embedding space where textual semantic information is effectively encoded.

% 2. Limitations
However, their output topic taxonomy seems plausible by itself but often fails to match with the complete taxonomy designed by a human curator,
%does not match with its actual topic structure, 
because they rely on only the text corpus in an unsupervised manner.
To be specific, their quality (e.g., coverage and accuracy) highly depends on the number of sub-topic clusters (i.e., child nodes), which has to be manually controlled by a user.
In addition, it is sensitive to the topic imbalance in the document collection, which makes it difficult to find out minor topics.
In the absence of any information about the topic hierarchy, the unsupervised methods intrinsically become vulnerable to these problems.

% 3. Topic hierarchy as weakly supervision
On the other hand, for some other text mining tasks or NLP applications, several recent studies have tried to take advantage of auxiliary information about the latent topic structure~\cite{meng2020discriminative, meng2020hierarchical, meng2019weakly, shen2021taxoclass, huang2020weakly, meng2020text, meng2018weakly}.
Most of them focus on utilizing a hierarchy of topic surface names as additional supervision, because it can be easily given as a user's interests or prior knowledge.
Specifically, they retrieve the top-$K$ relevant terms to each topic~\cite{meng2020discriminative, meng2020hierarchical} or train a hierarchical text classifier using unlabeled documents and the topic names~\cite{meng2019weakly, shen2021taxoclass}.
Despite their effectiveness, their major limitation is that they are only able to consider the known topics included in the given topic hierarchy.
That is, the coverage of the obtained results is strictly limited to the given topics.
Since it is very challenging for a user to be aware of a full topic structure, a naive solution to incorporate a user-provided hierarchy of topic names into the topic taxonomy is likely to only partially cover the text corpus.
%they cannot be directly used for constructing the topic taxonomy which should cover the entire text corpus and all the topics.
% Nevertheless, the weakly supervised methods can consider only the topics in the given hierarchy, which may not cover the entire text collection.
% In other words, their outputs are limited to the given topic hierarchy, and it results in vulnerability to incompleteness of the supervision.

% 4. New task introduction => Topic taxonomy completion
To tackle this limitation, we introduce a new problem setting, named topic taxonomy completion, to construct a complete topic taxonomy by making use of additional topic information assumed to be partial or incomplete. 
Formally, given a text corpus and its partial hierarchy of topic names, this task aims to identify the term clusters for each topic, while discovering the novel topics that do not exist in the given hierarchy but exist in the corpus.
Figure~\ref{fig:problem} illustrates a toy example of our task, where the novel topics (e.g, \textit{arts} and \textit{hockey}) are correctly detected and placed in the right position within the taxonomy.
This task can be practically applied not only for the case that a user's incomplete knowledge is available, but also for incremental management of the topic taxonomy.
In case that the document collection is constantly growing, and so are their topics, the out-dated topic taxonomy of the previous snapshot can serve as the partial hierarchy to capture emerging topics.

% 5. Challenges
The technical challenges of this task can be summarized as follows.
First, novel topics should be identified by considering the hierarchical semantic relationship among the topics.
In Figure~\ref{fig:problem}, the topic \textit{hockey} is not novel in terms of the root node, because it obviously belongs to its known sub-topic \textit{sports}.
However, \textit{hockey} should be detected as a novel sub-topic of \textit{sports} as it does not belong to any of the known sport sub-categories (i.e., \textit{soccer} and \textit{baseball}).
Second, the granularity of novel sub-topics and that of known sub-topics need to be kept similar with each other, to achieve the consistency of semantic specificity among sibling nodes.
In Figure~\ref{fig:problem}, the root node should insert a single novel sub-topic \textit{arts}, rather than two novel sub-topics \textit{music} and \textit{dance}, based on the semantic specificity of its known sub-topics (i.e., \textit{politics} and \textit{sports}).

% 6 . Proposed framework
In this work, we propose \proposed, a hierarchical topic discovery framework to complete the topic taxonomy by recursively identifying novel sub-topic clusters of terms.
For each topic node, \proposed performs (i) text embedding and (ii) text clustering, to assign the terms into one of either the existing child nodes (i.e., known sub-topics) or newly-created child nodes (i.e., novel sub-topics).
It first optimizes \textit{locally discriminative embedding} which enforces the discrimination among the known sub-topics~\cite{meng2020hierarchical, meng2020discriminative} by using the given topic surface names;
this helps to make a clear distinction between known and novel sub-topic clusters as well.
Then, it performs \textit{novelty adaptive clustering} which separately finds the clusters on novel-topic terms and known-topic terms, respectively.
In particular, \proposed selectively assigns the terms into the child nodes, referred to as \textit{anchor terms}, while filtering out general terms based on their semantic relevance and representativeness.
%Notably, it automatically determines the number of novel sub-topics, by selecting the optimal number that maximizes the consistency with the known sub-topics.

% 7. Experiments
Extensive experiments on real-world datasets demonstrate that \proposed successfully completes a topic taxonomy with missing (i.e., novel) topic nodes correctly inserted.
Our human evaluation quantitatively validates the superiority of topic taxonomies generated by \proposed, in terms of the topic coverage as well as semantic coherence among the topic terms.
%(i.e., the consistency with the ground-truth topic hierarchy)
Furthermore, \proposed achieves the best performance among all baseline methods for a downstream task, which trains a weakly supervised text classifier by using the topic taxonomy instead of document-level labels.
% For reproducibility, the implementation will be publicly available through the anonymized github repository during the review process.\footnote{https://github.com/taxocom-submission/taxocom}

\section{Related Work}
\label{sec:related}
% In this section, we briefly review the literature on (i) topic taxonomy construction to find out the latent topic structure of a text corpus, (ii) entity-level taxonomy (or set) expansion to discover new entity pairs in the hypernym-hyponym relation, and (iii) novelty detection for text data to determine whether each term or document belongs to given (or inlier) topics or not.

\smallsection{Topic Taxonomy Construction}
\label{subsec:taxogen}
Early work on hierarchical topic discovery mainly focused on generative probabilistic topic models, such as hierarchical Latent Dirichlet Allocation (\hlda)~\cite{blei2003hierarchical} and hierarchical Pachinko Allocation Model (hPAM)~\cite{mimno2007mixtures}.
They describe the topic hierarchy in a generative process and then estimate parameters by using inference algorithms, including variational Bayesian inference~\cite{blei2003latent} and collapsed Gibbs sampling~\cite{griffiths2004finding}.
With the advances in text embedding techniques, several recent studies started to employ hierarchical clustering methods on a term embedding space, where textual semantic information is effectively captured. 
By doing so, they can construct a \textit{topic taxonomy} whose node corresponds to a term cluster representing a single topic.
Specifically, to find out hierarchical topic clusters, \taxogen~\cite{zhang2018taxogen} recursively performed text embedding and clustering for each sub-topic cluster, and \nettaxo~\cite{shang2020nettaxo} additionally leveraged network-motifs extracted from text-rich networks.
However, since all of them are unsupervised methods that primarily utilize the input text corpus, the high-level architecture of their output taxonomies does not usually match well with the one designed by a human.

% To improve the quality and accuracy of the topic taxonomy, there have been several attempts to incorporate a user's prior knowledge about the latent topic structure;
% it is usually provided as a hierarchy of topic surface names or that of topic-specific keywords.
% Using the given hierarchy as the minimum guidance (i.e., weak supervision), they tried to mine the set of representative terms for each topic~\cite{meng2020hierarchical} or classify each document into one of the topics~\cite{meng2019weakly, shen2021taxoclass}.
% Nevertheless, the weakly supervised methods can consider only the topics in the given hierarchy, which may not cover the entire text collection.
% In other words, their outputs are limited to the given topic hierarchy, and it results in vulnerability to incompleteness of the supervision.
% For this reason, they are employed for mining the narrow topics that a user has prior knowledge about or be interested in, rather than identifying the full topic taxonomy.

\smallsection{Entity Taxonomy Expansion}
\label{subsec:taxoexpan}
Recently, there have been several attempts to construct the entity (or term-level) taxonomy from a text corpus by expanding a given seed taxonomy~\cite{shen2017setexpan,shen2018hiexpan,shen2020taxoexpan,huang2020corel,zeng2021enhancing,mao2020octet,yu2020steam}.
Note that the main difference of an entity taxonomy from a topic (or cluster-level) taxonomy is that its node represents a single entity or term, so it mainly focuses on the entity-level semantic relationships.
They basically discover new entities that need to be inserted into the taxonomy, by learning the ``is-a'' (i.e., hypernym-hyponym) relation of parent-child entity pairs in the seed taxonomy.
To infer the ``is-a''  relation of an input entity pair, they train a relation classifier based on entity embeddings~\cite{shen2018hiexpan}, a pre-trained language model~\cite{huang2020corel}, and graph neural networks (GNNs)~\cite{shen2020taxoexpan}.
%or additionally optimize the entity decoder to directly generate the new entity at a target position~\cite{zeng2021enhancing}.
%They mainly bootstrap the seed entities by the help of high-quality patterns or contexts extracted from the corpus~\cite{shen2017setexpan, huang2020guiding}.
Despite their effectiveness, the entity taxonomy cannot either show the semantic relationships among high-level concepts (i.e., topics or term clusters) or capture term co-occurrences in the documents; this makes its nodes difficult to correspond to the topic classes of documents. 
%Despite their effectiveness, they only focus on the entity-level semantic coherence within each cluster, without considering the intrinsic topic of each document at all.
Therefore, they are not suitable for expanding the latent topic hierarchy of documents, rather be useful for enhancing a knowledge base.

\smallsection{Novelty Detection for Text Data}
\label{subsec:novelty}
Novelty (or outlier) detection for text data,\footnote{Both \textit{novelties} and \textit{outliers} are assumed to be semantically deviating in an input corpus, but the novelties can form a dense cluster whereas the outliers cannot.}
which aims to detect the documents that do not belong to any of the given (or inlier) topics, has been researched in a wide range of NLP applications.
They define the novel-ness (or outlier-ness) based on how far each document is located from semantic regions representing the normality.
To this end, most unsupervised detectors measure the local/global density~\cite{breunig2000lof, sathe2016subspace} or estimate the normal data distribution~\cite{zhuang2017identifying,ruff2019self,manolache2021date} in a low-dimensional text embedding space.
On the other hand, supervised/weakly supervised novelty detectors~\cite{hendrycks2020pretrained, lee2020multi, lee2021out, zeng2021adversarial, fouche2020mining} also have been developed to fully utilize the auxiliary information about the inlier topics.\footnote{The topic labels of training documents are available for a supervised setting~\cite{fouche2020mining, lee2020multi, hendrycks2020pretrained, zeng2021adversarial}, and only topic names are provided for a weakly supervised setting~\cite{lee2021out}.}
They further optimize the embedding space to be discriminative among the topics, so as to clearly determine whether a document belongs to each inlier topic.
However, none of them considers the hierarchical relationships among the inlier topics, 
which makes them ineffective to identify novel topics from a large text corpus having a topic hierarchy.
%\footnote{It is also known as \textit{anomaly detection}, where only normal texts of a single topic class are given for training~\cite{ruff2019self, manolache2021date}, or \textit{out-of-domain detection}, where only in-domain texts of multiple topic classes are given for training~\cite{hendrycks2020pretrained}.}

\section{Problem Formulation}
\label{sec:problem}
% In this section, we formally describe our new problem and notations, and clarify its difference from the previous problem setting.

\subsection{Concept Definition}
\label{subsec:definition}

\begin{definition}[Topic taxonomy]
A \textit{topic taxonomy} $\taxo$ refers to a tree structure about the latent topic hierarchy of terms $\termset$ and documents $\docuset$.
Each node $c\in\taxo$ is described by a cluster of terms $\topicterms{c} (\subset \termset)$ representing a single conceptual topic.
The most representative term for the node becomes a \textit{center term} $t_{c}\in\topicterms{c}$, usually regarded as the topic surface name.
The child nodes of each topic node correspond to its sub-topics.\footnote{The terms ``child nodes'' and ``sub-topics'' are used interchangeably in this paper.}
For each node $c$, the set of its $K_c$ child nodes is denoted by $\topicchilds{c} (\subset\taxo) = \{s_{c,1}, \ldots, s_{c,K_c}\}$.
\end{definition}

\subsection{Topic Taxonomy Completion}
\label{subsec:problem}
%Based on the definition in Section~\ref{subsec:definition}, we introduce our target task.
\begin{definition}[Topic taxonomy completion]
The inputs are a text corpus $\docuset$, its term set $\termset$,\footnote{This term set can be automatically extracted from the input text corpus.}
and a partial hierarchy $\taxo^0$ of topic surface names.\footnote{This problem setting presumes that a single representative term of a topic node (e.g., topic name) can be easily given as minimum guidance to complete the topic taxonomy.}
The goal of topic taxonomy completion is to complete the topic taxonomy $\taxo(\supset\taxo^0)$ so that it can cover the entire topic structure of the corpus, being guided by the given topic hierarchy.
For each node in the taxonomy $c\in\taxo$, it finds out the set of topic terms $\topicterms{c}$ that are semantically coherent.
In other words, the given topic hierarchy is extended into a larger one by identifying and inserting new topic nodes, while allocating each term into either one of the existing nodes ($\in\taxo^0$) or newly-created nodes ($\in\taxo\textbackslash\taxo^0$).
%, where $K_c$ is the number of its child nodes.
\end{definition}

Figure~\ref{fig:problem} shows an example of topic taxonomy completion for a news corpus.
Similar to unsupervised topic taxonomy construction~\cite{zhang2018taxogen, shang2020nettaxo}, our task works on the set of unlabeled documents whose topic information (e.g., topic class label) is not available.
The main difference is that a partial topic hierarchy is additionally provided, which can be a user's incomplete prior knowledge or an out-dated topic taxonomy of a growing text collection.
From the perspective that the given hierarchy serves as auxiliary supervision for discovering the entire topic structure, this task can be categorized as a \textit{weakly supervised} hierarchical topic discovery.

\section{\proposed: Proposed Framework}
\label{sec:method}
% In this section, we propose \proposed, a topic modeling framework that expands a given topic hierarchy into the complete topic taxonomy.
% We first introduce the overview of \proposed, then present the details of its embedding and clustering procedures;
% for each topic node, \proposed assigns each term and document into one of the sub-topic nodes while discovering novel sub-topic clusters.

\begin{figure*}[t]
    \centering
    \includegraphics[width=\linewidth]{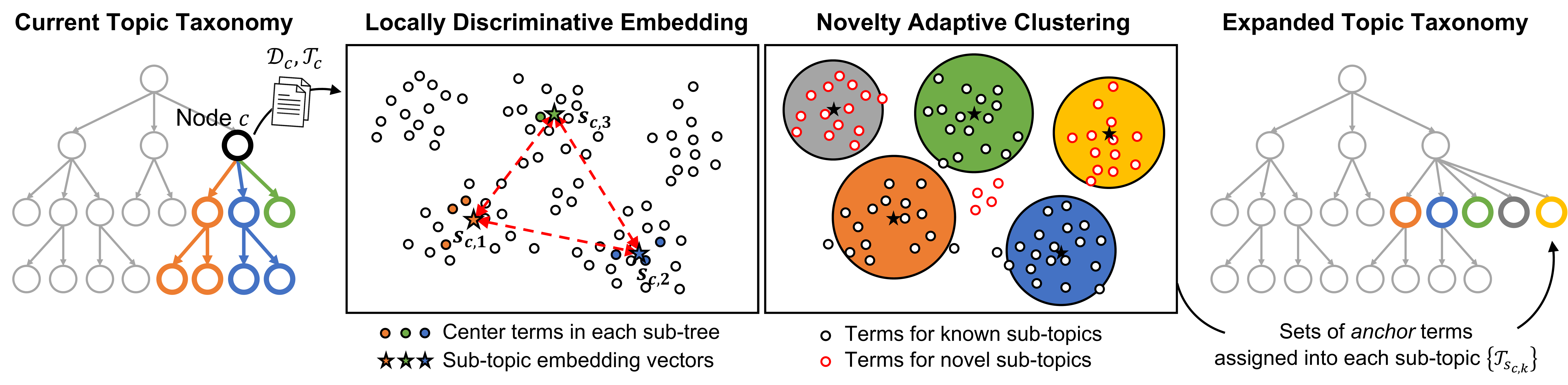}
    \caption{The overview of the \proposed framework which discovers the complete topic taxonomy by the recursive expansion of the given topic hierarchy. Starting from the root node, it performs (i) locally discriminative embedding and (ii) novelty adaptive clustering, to selectively assign the terms (of each node) into one of the child nodes. Best viewed in color.}
    \label{fig:framework}
\end{figure*}

\subsection{Overview}
\label{subsec:overview}
The proposed \proposed framework recursively expands the given hierarchy in a top-down approach.
Starting from the root node, \proposed performs (i) text embedding and (ii) text clustering for each node, to find out sub-topic clusters corresponding to its child nodes.
The key challenge here is to identify the term clusters for \textit{known} (i.e., given) sub-topics as well as \textit{novel} sub-topics, which cannot be covered by any of the known sub-topics, by leveraging the initial hierarchy as weak supervision.
\begin{itemize}
    \item \textbf{Locally discriminative embedding}: 
    \proposed optimizes the embedding space for the terms assigned to the current node $\topicterms{c} (\subset \termset)$.
    %by using the local text corpus that only contains the documents relevant to $\topicterms{c}$.
    Utilizing the topic surface names in the given hierarchy, the term embedding vectors are enforced to be discriminative among the known sub-topics, 
    so as to effectively compute the sub-topic membership of each term.
    
    \item \textbf{Novelty adaptive clustering}: 
    By using the term embeddings, 
    %\proposed finds out multiple term clusters of both known and novel sub-topics. 
    \proposed determines whether each term belongs to the known sub-topics or not based on its sub-topic membership.
    Then, it performs clustering which assigns each term into either one of the known sub-topics or novel sub-topics.
\end{itemize}
The taxonomy is expanded by inserting the obtained novel sub-topic clusters as the child nodes of the current node.
To fully utilize the documents relevant to $\topicterms{c}$ for both embedding and clustering, \proposed also produces the sub-corpus for each topic node $\topicdocs{c}(\subset\docuset)$ by assigning the documents into one of the sub-topics.\footnote{Although a document can cover multiple topics, we assign it into the most relevant sub-topic to exploit textual information from the topic sub-corpus during the process.}
%In addition, the set of anchor terms for each sub-topic is retrieved to recursively perform this process for its child nodes.
Figure~\ref{fig:framework} illustrates the overall process of \proposed.

\subsection{Locally Discriminative Text Embedding}
\label{subsec:embedding}
The goal of the embedding step is to obtain the low-dimensional embedding space that effectively encodes the textual similarity (or distance) among the terms assigned to a target topic node.
%to identify sub-topic clusters on the embedding space where
However, as pointed out in~\cite{gui2018expert, zhang2018taxogen, shang2020nettaxo}, the global embedding space trained on the entire text corpus is not good at capturing the distinctiveness among the topics, especially for lower-level topics.
For this reason, \proposed adopts the two strategies:
(i) \textit{local embedding}~\cite{gui2018expert}, which uses the subset of the entire corpus containing only the documents relevant to a target topic (e.g., the documents assigned to a specific topic node), and
(ii) \textit{keyword-guided discriminative embedding}~\cite{meng2020discriminative, meng2020hierarchical}, which additionally minimizes the semantic correlation among the pre-defined topics by utilizing their keyword sets.
%Since we have a prior knowledge of the sub-topics (i.e., child nodes) of the target node, we make the term vectors discriminative with each other,

\subsubsection{Local embedding}
\label{subsubsec:locemb}
To enhance the discriminative power of term embeddings at lower levels of the taxonomy by effectively capturing their finer-grained semantic information, \proposed employs the local embedding~\cite{gui2018expert} that uses the sub-corpus only relevant to the current topic $c$ instead of the entire corpus.
The most straightforward way to get the sub-corpus is simply using the set of documents assigned to the topic $c$, denoted by $\topicdocs{c}(\subset\docuset)$.
For the lower-level topics, however, it is more likely to include a small number of documents, which are not enough to accurately tune the embeddings.
For this reason, \proposed retrieves more relevant documents and uses them together with $\topicdocs{c}$.
Using the center term embedding of the topic $c$ as a query, it retrieves the top-$M$ closest terms and collects all the documents containing the terms.

\subsubsection{Keyword-guided discriminative embedding}
\label{subsubsec:disemb}
\proposed basically employs a spherical text embedding framework~\cite{meng2019spherical} to directly capture the semantic coherence among the terms into the directional (i.e., cosine) similarity in the spherical space.
Compared to other text embedding techniques learned in the Euclidean space~\cite{mikolov2013distributed, bojanowski2017enriching, pennington2014glove}, the spherical embedding is particularly effective for term clustering and similarity search, because it eliminates the discrepancy between its training procedure and practical usage.

From the given topic hierarchy, \proposed first builds the sub-topic keyword sets $\{\mathcal{K}_{s_{c,k}}\mid s_{c,k}\in\topicchilds{c}\}$ for the current node $c$, in order to use them as weak supervision for guiding the discrimination among the sub-topics. 
Each keyword set $\mathcal{K}_{s_{c,k}}$ collects all the center terms in the sub-tree rooted at the sub-topic node $s_{c,k}$.
For example, in Figure~\ref{fig:framework}, the center terms of sub-tree nodes (colored in orange, blue, and green, respectively) become the keywords of each sub-topic.
Since all topic names covered by each sub-tree surely belong to the corresponding sub-topic, they can serve as the sub-topic keywords for optimizing the discriminative embedding space.

The main objective of our text embedding is to maximize the probability $p(t_j| t_i)$ for the pairs of a term $t_i$ and its context term $t_j$ co-occurring in a local context window.
To model the generative likelihood of terms conditioned on each sub-topic $s_{c,k}$, it also maximizes $p(t|s_{c,k})$ for all terms in its keyword set $t\in\mathcal{K}_{s_{c,k}}$.
In addition, it makes the topic-conditional likelihood distribution $p(t|s_{c,k})$ clearly separable from each other, by minimizing the semantic correlation among the sub-topics $p(s_{c,j}|s_{c,i})$.
To sum up, the loss function for the node $c$ is described as follows.
\begin{equation}
\label{eq:josdloss}
\begin{split}
    \mathcal{L}_{emb} = &-\log \prod_{d\in\topicdocs{c}} \prod_{t_i\in d} \prod_{t_j\in\text{cw}(t_i;d)} p(t_j|t_i) \\
    &-\log \prod_{s_{c,k}\in\topicchilds{c}}\prod_{t_i\in\mathcal{K}_{s_{c,k}}} p(t_i|s_{c,k}) +\log \prod_{s_{c,i},s_{c,j}\in\topicchilds{c}} p(s_{c,j}|s_{c,i}),
\end{split}
\raisetag{48pt}
\end{equation}
where $\text{cw}(t_i;d)$ is the set of surrounding terms included in a local context window for a term $t_i$ in a document $d$.

Each probability (or likelihood) in Equation~\eqref{eq:josdloss} is modeled by using the embedding vector of each term $t_i$ and sub-topic $s_{c,k}$ (denoted by boldface letters $\tvec{i}$ and $\svec{c,k}$, respectively).
$p(t_i|s_{c,k})$ is defined by the von Mises-Fisher (vMF) distribution, which is a spherical distribution centered around the sub-topic embedding vector $\svec{c, k}$.
\begin{equation}
\label{eq:vmf}
    %p(d_i|c_k) &= \text{vMF}(\dvec{i};\cvec{k},\kappa_{c_k}) = n(\kappa_{c_k})\exp(\kappa_{c_k}\cos(\dvec{i}, \cvec{k})) \\
    p(t_i|s_{c,k}) = \text{vMF}(\tvec{i};\svec{c,k},\kappa_{s_{c,k}}) = C_{s_{c,k}}\exp(\kappa_{s_{c,k}}\cos(\tvec{i}, \svec{c,k}))
\end{equation}
where $\kappa_{s_{c,k}}\geq 0$ is the \textit{concentration} parameter, $C_{s_{c,k}}$ is the normalization constant, and the \textit{mean direction} of each vMF distribution is modeled by the sub-topic embedding vector $\svec{c,k}$.
The probability of term co-occurrence, $p(t_j|t_i)$, as well as that of inter-topic correlation, $p(s_{c,j}|s_{c,i})$ is simply defined by using the cosine similarity, % between their embedding vectors,
i.e., $p(t_j|t_i)\propto\exp(\cos(\tvec{i},\tvec{j}))$ and $p(s_{c,j}|s_{c,i})\propto\exp(\cos(\svec{c,i},\svec{c,j}))$.

Combining a max-margin loss function~\cite{vilnis2015word, vendrov2016order, ganea2018hyperbolic, meng2019spherical} with the probability (or likelihood) defined above, the objective of our text embedding in Equation~\eqref{eq:josdloss} is summarized as follows.
\begin{equation}
\label{eq:josdopt}
\begin{split}
    &\sum_{d\in\topicdocs{c}} \sum_{t_i\in d}\hspace{-18pt} \sum_{\qquad t_j\in\text{cw}(t_i;d)} \hspace{-12pt}  \left[\tvec{i}{}^\top\vvec{j'} - \tvec{i}{}^\top\vvec{j} + m \right]_+ + \hspace{-4pt}\sum_{s_{c,i},s_{c,j}\in\topicchilds{c}}\hspace{-4pt}\left[\svec{c,i}{}^\top\svec{c,j} - m\right]_+ \\
    &\quad - \sum_{s_{c,k}\in\topicchilds{c}}\sum_{t_i\in\mathcal{K}_{s_{c,k}}} \left(\log(C_{s_{c,k}}) + \kappa_{s_{c,k}}\tvec{i}{}^\top\svec{c,k}\right) \cdot \mathbbm{1}\left[\tvec{i}{}^\top\svec{c,k} < m\right]\\
    &\qquad \text{s.t.}\quad \forall t\in\topicterms{c}, s\in\topicchilds{c}, \quad \lVert\tvec{ }\rVert=\lVert\vvec{ }\rVert=\lVert\svec{ }\rVert=1, \kappa_c \geq 0,
\end{split}
\raisetag{12pt}
\end{equation}
where $[z]_+=\max(z,0)$ and $m$ is the margin size.
%, and $\mathbbm{1}$ is the indicator function.
Similar to previous work on text embedding~\cite{mikolov2013distributed, pennington2014glove}, a term $t_i$ has two independent embedding vectors as the target term $\tvec{i}$ and the context term $\vvec{i}$, and the negative terms $t_{j'}$ are randomly selected from the vocabulary.

% To sum up, the first term optimizes the similarity of each document and its words, and each word and its context words.
% The second term pulls the topic-indicative words (i.e., the given topic names) close to the corresponding topic vectors, while the third term makes the topic vectors far apart from each other.

\subsection{Novelty Adaptive Text Clustering}
\label{subsec:clustering}
\subsubsection{Novel-topic term identification}
\label{subsubsec:novelty}
The first step for novelty adaptive clustering is to determine whether each term can be assigned to one of the given sub-topics or not.
In other words, it distinguishes the terms that are relevant to the given sub-topics, referred to as \textit{known-topic terms} $\topickterms{c}$, from the terms that cannot be covered by them, referred to as \textit{novel-topic terms} $\topicnterms{c}$.
Since the vMF distributions for the given sub-topics are already modeled in the embedding space (Section~\ref{subsec:embedding}), they can be utilized for computing the confidence score that indicates how confidently each term belongs to one of the given sub-topics.
Specifically, the sub-topic membership of a term is defined by the softmax probability of its distance from each vMF distribution (i.e., sub-topic embedding vector), and the maximum softmax value is used as the confidence~\cite{hendrycks2020pretrained, lee2020multi}.
In the end, the novelty score of each term is defined as follows.
\begin{equation}
\label{eq:novscore}
    score_{nov}(t) = 1 - \max_{s_{c,k}\in\topicchilds{c}} \frac{\exp( \cos(\tvec{}, \svec{c,k})/T)} {\sum_{s_{c,k'}\in\topicchilds{c}} \exp(\cos(\tvec{}, \svec{c,k'}))/T)},
\end{equation}
where $T$ is the temperature parameter that controls the sharpness of the distribution.
Using the novelty threshold $\tau_{nov}$, the terms for the node $c$ is divided into the set of known-topic terms and novel-topic terms according to their novelty score.
\begin{equation}
\begin{split}
    \topickterms{c} &= \left\{t \mid score_{nov}(t) < \tau_{nov}, t\in\topicterms{c}\right\} \\
    \topicnterms{c} &= \left\{t \mid score_{nov}(t) \geq \tau_{nov}, t\in\topicterms{c}\right\}
\end{split}
\end{equation}
 
The novelty threshold is determined based on the number of the sub-topics $K_c$, i.e., $\tau_{nov} = \left(1-{1}/{K_c} \right)^\beta$ where $\beta\geq1$ is the hyperparameter to control the boundary of known-topic terms.
Note that a larger $\beta$ value incurs a smaller threshold value, which allows to identify a larger number of novel-topic terms, and vice versa.
The novelty score ranges in $(0, 1-{1}/{K_c}]$ because of the softmax on similarities with $K_c$ known sub-topics in Equation~\eqref{eq:novscore}.

\subsubsection{Spherical term clustering}
\label{subsubsec:clustering}
\proposed discovers the term clusters which cover both the known and novel sub-topics.
Namely, it assigns each known-topic term $t\in\topickterms{c}$ into one of the existing sub-topics $\topicchilds{c}$, and simultaneously, assigns each novel-topic term $t\in\topicnterms{c}$ into one of the newly-identified novel sub-topics $\novelchilds{c}$.
Finally, it outputs the sub-topic cluster assignment variables $\{z_t\mid t\in\topicterms{c}\}$.
%and $z_d$ for all the terms $t\in\topicterms{c}$ and documents $d\in\topicdocs{c}$.

\smallsection{Known-topic term clustering}
In terms of known-topic terms $\topickterms{c}$, \proposed allocates each term into its closest known sub-topic in the embedding space;
i.e., $z_t = \argmax_{s_{c,k}\in\topicchilds{c}} \cos(\tvec{}, \svec{c, k})$.
%The embedding vector of each sub-topic $s_{c,k}$ is denoted by boldface letters $\svec{c, k}$.
%\footnote{Instead of the sub-topic vector $\svec{c, k}$, its center term vector $\tvec{s_{c,k}}$ also can be used.}
% \begin{equation}
% \label{eq:termassign}
%     z_t = \argmax_{s_{c,k}\in\topicchilds{c}} \ \cos(\tvec{}, \svec{c, k}).
% \end{equation}

\smallsection{Novel-topic term clustering}
Unlike known sub-topics whose embedding vector $\svec{c,k}$ and center term vector $\tvec{s_{c,k}}$ are available for clustering, there does not exist any information about novel sub-topics.
For this reason, \proposed performs $K_c^*$-means spherical clustering~\cite{dhillon2001concept} on the novel-topic terms $\topicnterms{c}$, thereby obtaining the mean vector and center term of each cluster.\footnote{We also considered density-based clustering (e.g., DBSCAN~\cite{campello2015hierarchical}) for identifying novel sub-topics, but we empirically found that it is quite sensitive to hyperparameters as well as cannot consider the semantic relevance to the center terms of each sub-topic.}
The number of novel sub-topic clusters $K_c^*$ is determined to balance the semantic specificity among clusters, which will be discussed in Section~\ref{subsubsec:movmf}.
%One remaining challenge is to select the number of the clusters for spherical clustering, and we will discuss about it in Section~\ref{subsubsec:movmf}.
%For the node $c$, the set of novel sub-topics is denoted by $\novelchilds{c}$.

\subsubsection{Anchor term selection}
\label{subsubsec:anchorterm}
The initial term clustering results from Section~\ref{subsubsec:clustering} contain the cluster assignment variable of all the terms, but not every term of a topic necessarily belongs to one of its sub-topics.
For example, the term \textit{game} in the topic node \textit{sports} does not belong to any of its child nodes, representing specific sport sub-categories, such as \textit{tennis}, \textit{baseball}, and \textit{soccer}.
Thus, it is necessary to carefully mine the set of anchor terms, which are apparently relevant to each sub-topic $s_{c,k}$.
%, and filter out non-anchor terms to their parent node $c$.

To this end, \proposed defines the significance score of a term by considering both its \textit{semantic relevance} to each sub-topic cluster, denoted by $rel(t, s_{c,k})$, and the \textit{representativeness} in the corresponding sub-corpus $\topicdocs{s_{c,k}}$, denoted by $rep(t, s_{c,k})$.
\begin{equation}
\label{eq:sigscore}
    score_{sig}(t) = \max_{s_{c,k}\in\topicchilds{c}\cup\novelchilds{c}} \left[rel(t, s_{c,k}) \times rep(t, s_{c,k})\right].
\end{equation}
To be specific, the semantic relevance is computed by the cosine similarity between their embedding vectors, while the representativeness is obtained based on the term frequency in the sub-corpus.
By doing so, it can make use of both information from the embedding space and the term occurrences.
\begin{equation}
\begin{split}
\label{eq:relrepscore}
    rel(t, s_{c,k}) &= \cos(\mathbf{t}, \svec{c,k}) \\
    rep(t, s_{c,k}) &= \left(int(t, s_{c,k}) \times dis(t, s_{c,k}) \times pop(t, s_{c,k}) \right)^{1/3}.
\end{split}
\end{equation}

%Next, to obtain the sub-corpus for each sub-topic \proposed assigns each document into one of the sub-topics $\topicchilds{c}\cup\novelchilds{c}$, by aggregating the cluster assignment of its terms based on the tf-idf weights.
For mining the representativeness from the sub-corpus $\topicdocs{s_{c,k}}$, \proposed collects the documents by aggregating the cluster assignment of their terms based on the tf-idf weights.
That is, a document chooses its sub-topic cluster based on how many terms are assigned to each sub-topic cluster considering their importance as well.
The cluster assignment of a document $z_d$ is defined as follows.
%is described as follows.
%obtained by using the cluster-specific tf-idf score based on the term-cluster assignment $z_t$.
\begin{equation}
\label{eq:docassign}
    z_d = \argmax_{s_{c,k}\in\topicchilds{c}\cup\novelchilds{c}} \sum_{t\in d} \mathbb{I}[z_t==s_{c,k}]\cdot \text{tf}(t, d) \cdot \text{idf}(t).
\end{equation}
Motivated by context-aware semantic online analytical processing (\caseolap)~\cite{tao2016multi},
the representativeness is defined as a function of three criteria:
(i) \textit{Integrity} -- A term with high integrity refers to a meaningful and understandable concept.
This score can be simply calculated by the state-of-the-art phrase mining technique, such as \segphrase~\cite{liu2015mining} and \autophrase~\cite{shang2018automated}.
(ii) \textit{Distinctiveness} -- A distinctive term has relatively strong relevance to the sub-corpus of the target sub-topic, distinguished from its relevance to other sub-corpora of sibling sub-topics.
The distinctiveness score is defined by using the BM25 relevance measure,
$dis(t, s_{c,k})$ $={\exp (\text{BM25}(t, \topicdocs{s_{c,k}}))}/$ ${(1 + \sum_{s_{c,k'}\in\topicchilds{c}\cup\novelchilds{c}}\exp (\text{BM25}(t, \topicdocs{s_{c,k'}})))}$.
(iii) \textit{Popularity} -- A term with a high popularity score appears more frequently in the sub-corpus of the target sub-topic than the others, $pop(t, s_{c,k}) = {\log ( \text{tf}(t, \topicdocs{s_{c,k}}) + 1)}/{\log (\sum_{t'\in\topicterms{c}}\text{tf}(t', \topicdocs{s_{c,k}}))}$.
% \begin{equation}
% \begin{split}
%     dis(t, s_{c,k}) &=\frac{\exp (BM25(t, \topicdocs{s_{c,k}}))}{1 + \sum_{s_{c,k'}\in\topicchilds{c}\cup\novelchilds{c}}\exp (BM25(t, \topicdocs{s_{c,k'}}))} \\
%     pop(t, s_{c,k}) &=\frac{\log ( \text{tf}(t, \topicdocs{s_{c,k}}) + 1)}{\log (\sum_{t'\in\topicterms{c}}\text{tf}(t', \topicdocs{s_{c,k}}))}.
% \end{split}
% \end{equation}

Finally, \proposed only keeps the anchor terms whose significance score is larger than the threshold $\tau_{{sig}}$,
after filtering out the general terms that are less informative to represent each sub-topic.
%, which means that it is more general term and less likely to belong to one of sub-topics.
%Low-scored terms, indicating are not representative for any sub-topics, are considered as general terms and pushed back to the parent.
\begin{equation}
\begin{split}
\label{eq:termassign}
    \topicterms{s_{c,k}} &= \left\{t\mid z_t==s_{c,k}, score_{sig}(t)\geq\tau_{sig}, \forall t\in\topicterms{c}, \right\} \\
    %\topicdocs{s_{c,k}} &= \left\{d\mid z_d==s_{c,k}, \forall d\in\topicdocs{c}\right\}
\end{split}
\end{equation}

\subsubsection{Novel sub-topic cluster refinement}
\label{subsubsec:movmf}
Using the set of anchor terms, \proposed estimates the vMF distribution (i.e., mean vector and concentration parameter) for each sub-topic cluster in the embedding space.
This final step is designed to choose the proper number of novel clusters $K_c^*$ (in Section~\ref{subsubsec:clustering}), with the help of the estimated concentration values indicating the semantic specificity of each sub-topic cluster.
Formally, it selects the $K_c^*$ value so that it can minimize the standard deviation of all the concentrations, i.e., $\argmin_{K_c^*} \text{stdev}\left[\{\kappa_{s_{c,k}} \mid \forall s_{c,k}\in\topicchilds{c}\cup\novelchilds{c}\}\right]$.
In this process, to measure the standard deviation based on the identified novel sub-topics $\novelchilds{c}$ for each $K_c^*$ value, a part of the clustering step (from Section~\ref{subsubsec:clustering} to~\ref{subsubsec:movmf}) are repeated.
Notably, \proposed is capable of automatically finding the total number of sub-topics, by harmonizing the semantic specificity of novel sub-topics with that of known sub-topics, whereas unsupervised methods for topic taxonomy construction rely on a user's manual selection.

\section{Experiments}
\label{sec:exp}
% In this section, we present the experimental results that demonstrate the effectiveness of the \proposed framework.
% We first quantitatively evaluate the topic assignment for documents by using their ground-truth topic class labels.
% Then, we also qualitatively  analyses on the output topic taxonomies obtained by hierarchical topic discovery methods.

\subsection{Experimental Settings}
\label{subsec:expset}
\subsubsection{Datasets}
% Copied from OOCD paper
For our experiments, we use two real-world datasets collected from different domains, \textbf{\nyt}\footnote{The news articles are crawled by using https://developer.nytimes.com/} and \textbf{\arxiv}\footnote{The abstracts of \arxiv papers are crawled from https://arxiv.org/}, and they have a two-level hierarchy of topic classes.
Thus, we regard the hierarchies as the ground-truth provided by a human curator, and use them to evaluate the completeness of topic taxonomies.
%The documents were crawled from 26 categories in 5 different sections (for \nyt), and 34 categories in 3 different sections (for \arxiv).
For both the datasets, the number of documents for each topic class is not balanced, and AutoPhrase~\cite{shang2018automated} is used to tokenize raw texts of each document.
The statistics are summarized in Table~\ref{tbl:datastats}.

\begin{table}[t]
\caption{The statistics of the datasets.}
\centering
\small
%\resizebox{0.99\linewidth}{!}{%
\begin{tabular}{c|cccc}
    \toprule
    Corpus & Avg-Length & \#Topics & \#Documents & \#Terms  \\\midrule
    \textbf{\nyt} & 739.8 & 5 $\rightarrow$ 26 & \ \ 13,081 & 23,245 \\
    \textbf{\arxiv} & 123.5 & 3 $\rightarrow$ 48 & 230,018 & 24,148 \\\bottomrule
\end{tabular}
%}
\label{tbl:datastats}
\end{table}

\begin{table}[t]
\caption{The topic classes deleted from the original topic hierarchy. ($\rightarrow\ast$) denotes all the sub-topics.}
\label{tbl:partialtaxo}
\centering
\resizebox{0.99\linewidth}{!}{%
\begin{tabular}{cll}
\toprule
 & {\textbf{\nyt}} & {\textbf{\arxiv}} \\\midrule
$\taxoa$ & \textit{arts}($\rightarrow\ast$) & \textit{physics}($\rightarrow\ast$) \\\midrule
$\taxob$ & \textit{arts}$\rightarrow$\textit{movies} & \textit{cs}$\rightarrow$\textit{AI,CV,DC,GT,IT} \\
         & \textit{business}$\rightarrow$\textit{stocks} & \textit{math}$\rightarrow$\textit{CA,DS,GR,RT} \\
         & \textit{politics}$\rightarrow$\textit{abortion,budget,insurance} & \textit{physics}$\rightarrow$\textit{chem-ph,gen-ph} \\
         & \textit{sports}$\rightarrow$\textit{baseball,hockey} & \hspace{33pt} \textit{plasm-ph} \\\midrule
$\taxoc$ & \textit{arts}$\rightarrow$\textit{music} & \textit{cs}($\rightarrow\ast$) \\ 
         & \textit{politics}$\rightarrow$\textit{gun control,military} & \textit{math}$\rightarrow$\textit{CO,DG,PR} \\
         & \textit{science}($\rightarrow\ast$) &\textit{physics}$\rightarrow$\textit{atom-ph,flu-dyn} \\
         & \textit{sports}$\rightarrow$\textit{football,soccer} & \\
\bottomrule
\end{tabular}
}
\end{table}

To investigate the effect of an initial topic hierarchy, we consider three scenarios using different partial topic hierarchies with different completeness.
Each partial hierarchy is generated by randomly deleting a few topics from the ground-truth topic hierarchy:
(i) $\taxoa$ deletes only a single first-level topic (and all of its second-level sub-topics), (ii) $\taxob$ drops some of the second-level topics, and (iii) $\taxoc$ does for both levels.
The deleted topics are listed in Table~\ref{tbl:partialtaxo}.

\subsubsection{Baseline Methods}
We consider several methods that are capable of constructing a topic taxonomy (or discovering hierarchical topics) as the baselines.
They can be categorized as either unsupervised methods using only an input corpus, or weakly supervised methods initiated with a given topic hierarchy.

\begin{itemize}
    \item \textbf{\hlda}~\cite{blei2003hierarchical}: Hierarchical latent Dirichlet allocation.
    The document generation process is modeled by selecting a path from the root to a leaf and sampling its words along the path. 
    
    \item \textbf{\taxogen}~\cite{zhang2018taxogen}: The unsupervised framework for topic taxonomy construction. 
    To identify term clusters, it employs the text embedding space, optimized by SkipGram~\cite{mikolov2013distributed}.
    %The number of child nodes is manually set, as done in~\cite{zhang2018taxogen,shang2020nettaxo}.
    
    % \item \textbf{\weshclass}~\cite{meng2019weakly}, \textbf{\taxoclass}~\cite{shen2021taxoclass}:  Hierarchical text classification methods that train their classifier in a weakly supervised setting.
    % The hierarchy of topic surface names is utilized to generate pseudo-documents~\cite{meng2019weakly} or pseudo-labels~\cite{shen2021taxoclass}, which are used for training the classifier.
    
    \item \textbf{\josh}~\cite{meng2020hierarchical}: Hierarchical text embedding technique to mine the set of relevant terms for each given topic. 
    It finds the topic terms based on the directional similarity.
    %During the optimization, it expands the set of topic terms based on the directional similarity, derived from the vMF distribution.
    
    \item \textbf{\corel}~\cite{huang2020corel}: Seed-guided entity taxonomy expansion method.
    It first expands only the topic names based on its relation classifier, then retrieves the topic terms based on the relevance in the embedding space induced by SkipGram~\cite{mikolov2013distributed}.
    
    \item \textbf{\proposed}: The proposed framework for topic taxonomy completion, which finds out novel sub-topic clusters to expand the topic taxonomy in a hierarchical manner.
\end{itemize}
Note that \corel directly learns the ``is-a'' relation of the entity (i.e., topic name) pairs in the given topic hierarchy to discover novel entity pairs.
On the contrary, \proposed implicitly infers the relation at the cluster-level based on its recursive clustering.
Furthermore, \corel mines the topic terms solely based on the embedding similarity, whereas \proposed additionally considers the representativeness in the sub-corpus relevant to the topic, as described in Equation~\eqref{eq:sigscore}.

% \subsubsection{Evaluation Scenarios}
% \label{subsubec:evalscenario}
% We use the Davies-Bouldin index (\dbi)~\cite{davies1979cluster} to assess the quality of clustering results~\cite{zhang2018taxogen}.
% It assign the maximum similarity scores as bla.
% We measure the normalized mutual information (NMI) score between the ground-truth topic labels and the obtained cluster indices.

\begin{table}[t]
\caption{Human evaluation on the output topic taxonomy.}
\centering
\resizebox{0.99\linewidth}{!}{%
\begin{tabular}{ccRRRR}
    \toprule
     \multirow{2.5}{*}{\shortstack{\textbf{Given}\\$\mathbf{\bm{\mathcal{H}^0}}$}} & \multirow{2.5}{*}{\shortstack{\textbf{Methods}}} & \multicolumn{2}{c}{\textbf{\nyt}} & \multicolumn{2}{c}{\textbf{\arxiv}} \\\cmidrule(lr){3-4}\cmidrule(lr){5-6}
    & & {\small\textbf{Coherency}} & {\small\textbf{Complete.}} & {\small\textbf{Coherency}} & {\small\textbf{Complete.}} \\\midrule
    \multirow{2}{*}{-} 
    & \hlda & 0.3033 & 0.3161 & 0.2211 & 0.3755 \\
    & \taxogen & 0.7406 & 0.5640 & 0.7644 & 0.5242 \\\midrule
    
    \multirow{3.2}{*}{$\taxoa$}
    & \josh & 0.8781 & 0.8387 & 0.8753 & 0.7843 \\
    & \corel & 0.8690 & 0.8452 & 0.8369 & 0.8458 \\
    & \proposed & \textbf{0.8811} & \textbf{0.9640} & \textbf{0.8913} & \textbf{0.9379} \\\midrule
    
    \multirow{3.2}{*}{$\taxob$}
    & \josh & 0.8583 & 0.7742 & 0.8364 & 0.7647 \\
    & \corel & 0.8431 & 0.9052 & 0.8429 & 0.8350 \\
    & \proposed & \textbf{0.8633} & \textbf{0.9303} & \textbf{0.8667} & \textbf{0.8556} \\\midrule
    
    \multirow{3.2}{*}{$\taxoc$} 
    & \josh & 0.8467 & 0.7419 & 0.8585 & 0.5882 \\
    & \corel & 0.8347 & 0.8426 & 0.8433 & 0.7046 \\
    & \proposed & \textbf{0.8556} & \textbf{0.9077} & \textbf{0.8613} & \textbf{0.8951} \\\bottomrule
\end{tabular}
}
\label{tbl:humaneval}
\end{table}

\subsection{Quantitative Evaluation}
\label{subsec:doccluster}
\subsubsection{Human evaluation on the quality of topic taxonomy}
\label{subsubsec:humaneval}
First, we assess the quality of topic taxonomies by using human domain knowledge. 
To this end, we recruit 10 doctoral researchers as evaluators, and ask them to perform two tasks that examine the following aspects of a topic taxonomy.\footnote{They are allowed to use web search engines when encountering unfamiliar terms.}
\textbf{(i) Term coherency} indicates how strongly the terms in a topic node are semantically coherent.
Similar to previous topic model evaluations~\cite{xie2015incorporating, shang2020nettaxo}, the top-10 terms of each topic node are presented to human evaluators, and they are requested to identify how many terms are relevant to their common topic (or center term).
The coherency is defined by the ratio of the number of relevant terms over the total number of presented terms.
\textbf{(ii) Topic completeness} quantifies how completely the set of topic nodes covers the ground-truth topics.
For each level, every topic name in the ground-truth topic hierarchy is given as a query, and the set of output topic nodes becomes the support set. 
Human evaluators are asked to rate the score $\in[0,1]$ how confidently the query belongs to one of the topics in the support set (i.e., similarity with the semantically closest support topic).
The completeness is defined by the average score for all the queries. 

In Table~\ref{tbl:humaneval}, \proposed significantly outperforms all the baselines in terms of both the measures.\footnote{We first test the inter-rater reliability on ranks of the methods. We obtain the Kendall coefficient of 0.96/0.91 (\nyt) and 0.94/0.89 (\arxiv) respectively for the coherence and completeness, which indicates the consistent assessment of the 10 evaluators.}
For topic completeness, the weakly supervised methods beat the unsupervised methods by a large margin, because their output topic taxonomy at least covers all the topics in the given topic hierarchy.
Notably, \proposed gets higher scores than \josh and \corel, which implies that it more accurately discovers ground-truth topics deleted from the full hierarchy.
In addition, \proposed is ranked first for the term coherency, as it captures each term's representativeness in the topic-relevant documents as well as its semantic relevance to the target cluster.

\subsubsection{Weakly supervised document classification using topic taxonomy}
\label{subsubsec:weshclass}
Next, we indirectly evaluate each output topic taxonomy by making use of a downstream task that takes a topic taxonomy as its input.
We compare the performance of \weshclass~\cite{meng2019weakly}, a weakly supervised hierarchical text classifier trained by using only unlabeled documents and the hierarchy of target classes (with the class-specific keywords), rather than document-level class labels.
Specifically, we use the topic taxonomy obtained by \proposed and the baseline methods as the hierarchy of target classes, and its top-10 topic terms serve as the class-specific keywords.
We measure the normalized mutual information (NMI) between predicted document topic labels and ground-truth ones in terms of clustering, as well as MacroF1 and MicroF1 in terms of classification.\footnote{In case of topic taxonomies generated by weakly supervised methods, to measure the F1 scores based on document-level topic class labels, we manually find the one-to-one mapping from identified novel topic nodes to the deleted ground-truth topic classes.}

Table~\ref{tbl:docclass} reports that \weshclass achieves the best NMI and F1 scores when being trained using the output topic taxonomy of \proposed.
The final classification performance of \weshclass is mainly affected by (i) the keyword (i.e., top-10 terms) coverage for each topic class and (ii) the topic coverage for the entire text corpus.
As analyzed in Section~\ref{subsubsec:humaneval}, \proposed generates more complete topic taxonomies compared to \josh and \corel, which helps \weshclass to learn the discriminative features for a larger number of ground-truth topic classes in each dataset. 
Besides, since the topic terms retrieved by \proposed captures additionally the representativeness in the topic-relevant documents, they become more informative and accurate class-specific keywords for training \weshclass, which eventually leads to better performances.
%\josh can only find topic terms of the known topics given in an initial topic hierarchy, so it is not able to predict the topic classes that do not exist in the given hierarchy.
%\corel by, and it fails to effectively find out missing topic nodes, which makes the final performance worse.
In conclusion, the higher quality topic taxonomy of \proposed can provide much more useful supervision for the downstream task on unlabeled documents.

\begin{table}[t]
\caption{Performance of the weakly supervised document classifier (i.e., \weshclass~\cite{meng2019weakly}), trained by using the output topic taxonomy of each method.}
\centering
\resizebox{0.99\linewidth}{!}{%
\begin{tabular}{ccQQQQQQ}
    \toprule
    \multirow{2.5}{*}{\shortstack{\textbf{Given}\\$\mathbf{\bm{\mathcal{H}^0}}$}} & \multirow{2.5}{*}{\shortstack{\textbf{Methods}}} & \multicolumn{3}{c}{\textbf{\nyt}} & \multicolumn{3}{c}{\textbf{\arxiv}} \\\cmidrule(lr){3-5}\cmidrule(lr){6-8}
    & & {\small \textbf{NMI}} & {\small \textbf{MacroF1}} & {\small \textbf{MicroF1}} & {\small \textbf{NMI}} & {\small \textbf{MacroF1}} & {\small \textbf{MicroF1}} \\\midrule
    \multirow{2}{*}{-}
    & \hlda & 0.4886 & - & - & 0.2348 & - & - \\
    & \taxogen & 0.7198  & - & - & 0.4307 & - & - \\\midrule
             & \josh  & 0.6815 & 0.4251 & 0.6334 & 0.3172 & 0.1115 & 0.1389  \\
    $\taxoa$ & \corel  & 0.7074 & 0.5036 & 0.6345 & 0.4141 & 0.2927 & 0.3301  \\
             & \proposed & \textbf{0.7630} & \textbf{0.6205} & \textbf{0.7980} & \textbf{0.4391} & \textbf{0.3494} & \textbf{0.3928} \\\midrule
             & \josh  & 0.6099 & 0.3086 & 0.4339 & 0.3753 &  0.1446 & 0.1767  \\
    $\taxob$ & \corel  & 0.6524 & 0.3958 & 0.5260 & 0.4449 & 0.3196 & 0.3702  \\
             & \proposed & \textbf{0.7520} & \textbf{0.5443} & \textbf{0.7738} & \textbf{0.4848} & \textbf{0.3795} & \textbf{0.4428} \\\midrule
             & \josh & 0.6972& 0.3661 & 0.5707 & 0.3222 & 0.1250 & 0.1430 \\
    $\taxoc$ & \corel  & 0.7413 & 0.4309 & 0.7576 & 0.4242 & 0.2703 & 0.3336 \\
             & \proposed & \textbf{0.7795} & \textbf{0.5856} & \textbf{0.8333} & \textbf{0.4577} & \textbf{0.3293} & \textbf{0.3937} \\\bottomrule
\end{tabular}
}
\label{tbl:docclass}
\end{table}

\begin{table}[t]
\caption{Performance for document-level novelty detection.}
\centering
\resizebox{0.99\linewidth}{!}{%
\begin{tabular}{cccQQQQQQ}
    \toprule
     \multirow{2.5}{*}{\shortstack{\textbf{Given}\\$\mathbf{\bm{\mathcal{H}^0}}$}} & \multirow{2.5}{*}{\textbf{LE}} & \multirow{2.5}{*}{\textbf{DE}} & \multicolumn{3}{c}{\textbf{\nyt}} & \multicolumn{3}{c}{\textbf{\arxiv}} \\\cmidrule(lr){4-6}\cmidrule(lr){7-9}
    & & & \textbf{P} & \textbf{R} & \textbf{F1} & \textbf{P} & \textbf{R} & \textbf{F1} \\\midrule
    \multirow{4}{*}{$\taxoa$} 
    & & & 0.5833 & 0.6483 & 0.6141 & 0.2962 & 0.6440 & 0.4058   \\
    & \checkmark & & 0.6355 & 0.6298 & 0.6326 & 0.4235 & 0.5668 & 0.4848  \\
    & & \checkmark & 0.7033 & 0.6924 & 0.6978 & 0.4715 & 0.6664 & 0.5523   \\
    & \checkmark & \checkmark & \textbf{0.7878} & \textbf{0.7176} & \textbf{0.7511} & \textbf{0.5510} & \textbf{0.8424} & \textbf{0.6662}  \\\midrule
    
    \multirow{4}{*}{$\taxob$} 
    & & & 0.7308 & 0.5719 & 0.6417 & 0.2743 & 0.5514 & 0.3664  \\
    & \checkmark & & 0.7508 & 0.5725 & 0.6496 & 0.2994 & \textbf{0.5824} & 0.3955  \\
    & & \checkmark & 0.8082 & 0.5806 & 0.6758 & 0.3679 & 0.5330 & 0.4353   \\
    & \checkmark & \checkmark & \textbf{0.8470} & \textbf{0.7852} & \textbf{0.8149} & \textbf{0.5399} & 0.5042 & \textbf{0.5214} \\\midrule
    
    \multirow{4}{*}{$\taxoc$} 
    & & & 0.4090 & 0.6930 & 0.5144 & 0.4723 & 0.8315 & 0.6024   \\
    & \checkmark & & 0.5318 & 0.7358 & 0.6174 & 0.4321 & 0.7925 & 0.5593 \\
    & & \checkmark & 0.4532 & 0.7210 & 0.5566 & 0.5721 & 0.8527 & 0.6848 \\
    & \checkmark & \checkmark & \textbf{0.8761} & \textbf{0.7798} & \textbf{0.8251} & \textbf{0.6222} & \textbf{0.8808} & \textbf{0.7293 } \\\bottomrule
    
\end{tabular}
}
\label{tbl:docdetection}
\end{table}

% \subsubsection{Evaluation on document-topic assignment}
% \label{subsubsec:doccluster}
% In case of the hierarchical topic discovery methods that obtain the document-level topic assignment variables as well (i.e., \hlda, \taxogen, and \proposed), we measure its consistency between the assignment and their actual topic labels.
% That is, we evaluate the topic discriminative power of the output topic taxonomy.
% Simlar to Section~\ref{subsubsec:weshclass}, we measure the clustering NMI and NMI*, each of which considers all the documents and. known-topic documents, respectively. 

\begin{figure}[t]
    \centering
    \includegraphics[width=\linewidth]{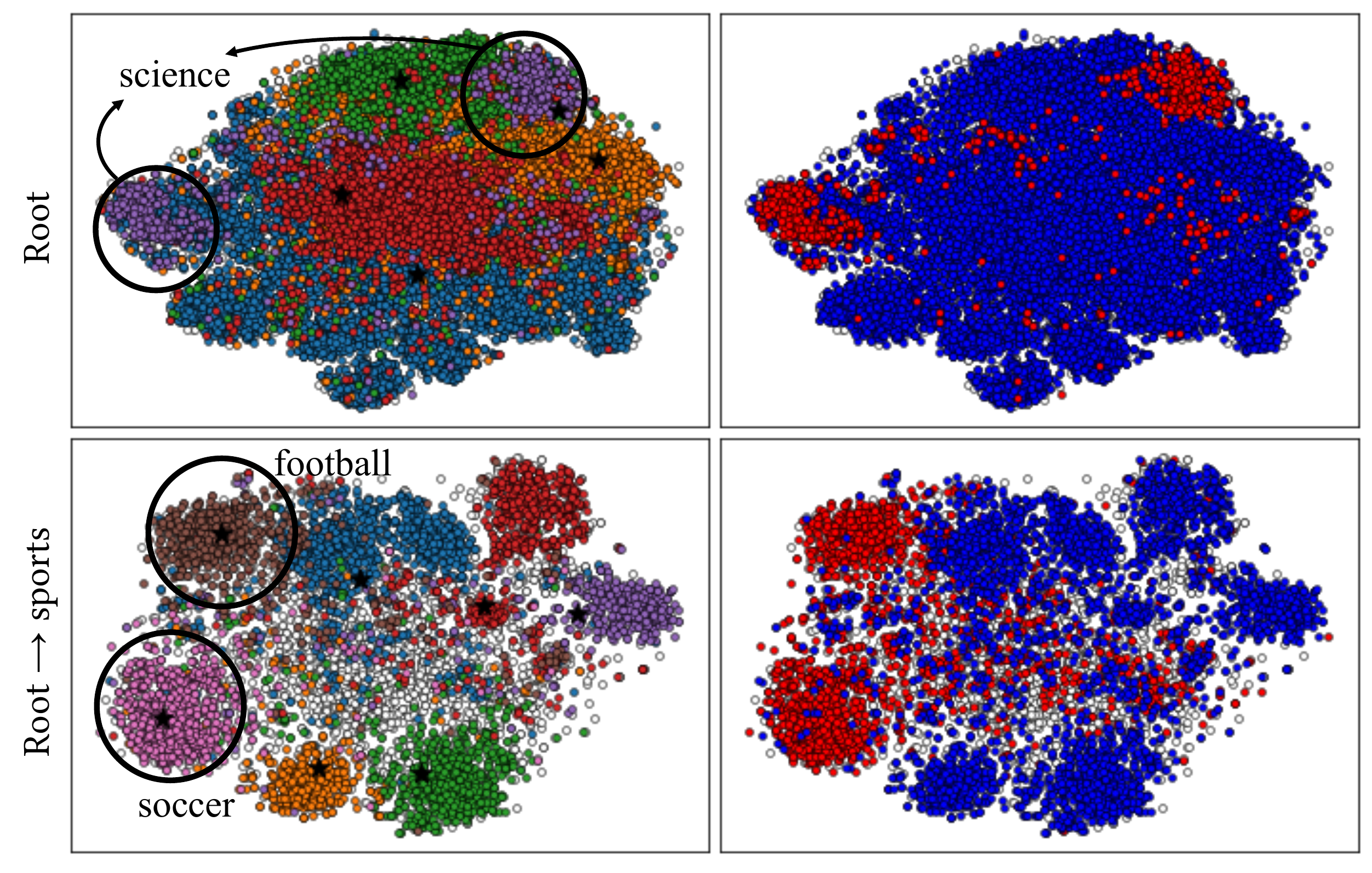}
    \caption{The embedding space of \proposed, which finds multiple sub-topic term clusters (Left) and discriminates between known-topic and novel-topic terms, respectively colored in blue and red (Right). Best viewed in color.}
    \label{fig:embspace}
\end{figure}

\begin{figure*}[t]
    \centering
    \includegraphics[width=\linewidth]{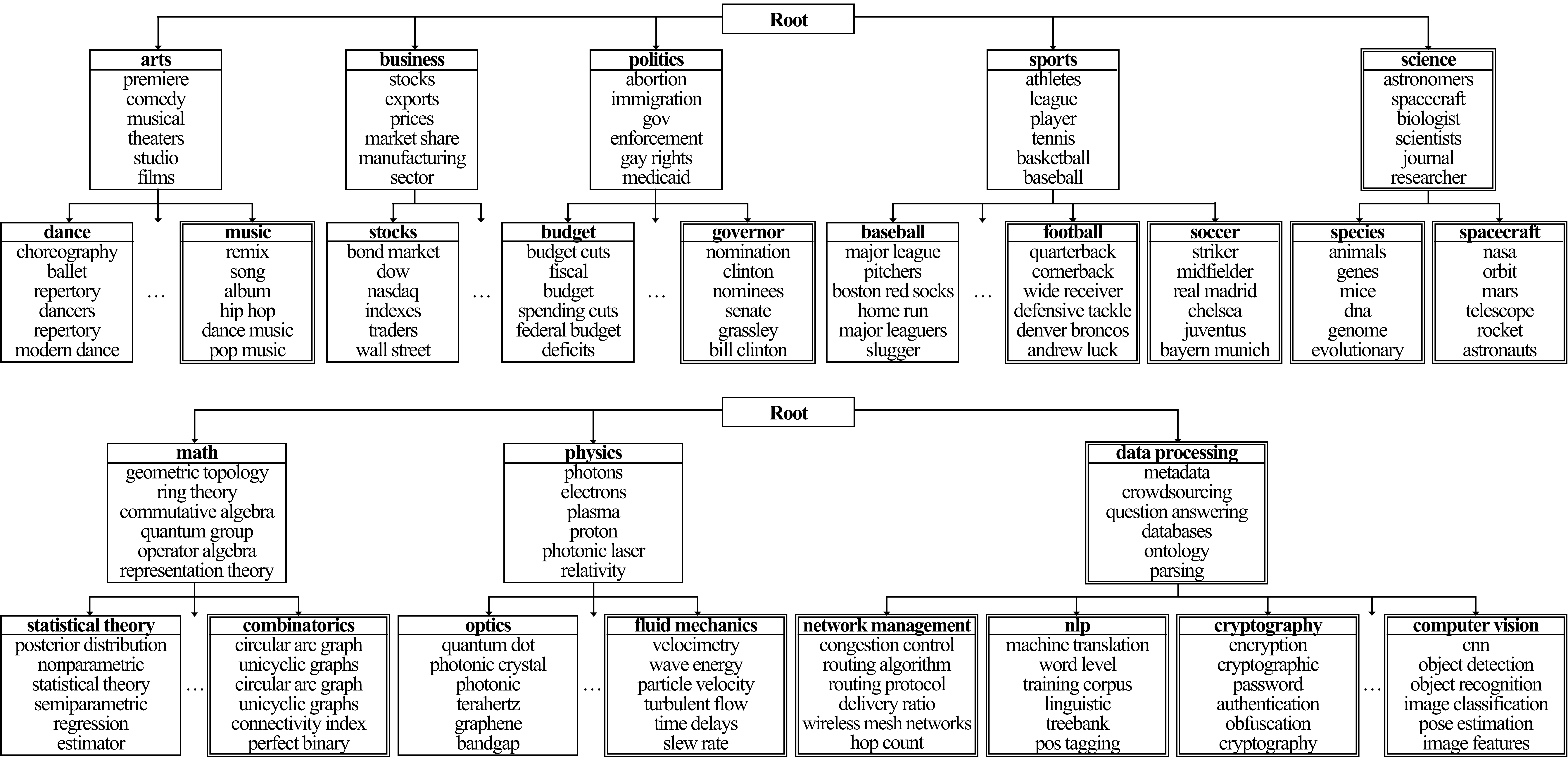}
    \caption{The output topic taxonomy of \proposed, where $\taxoc$ is given as the initial topic hierarchy. Double-lined boxes represent the newly inserted topic nodes. Dataset: \nyt (Upper) and \arxiv (Lower).}
    \label{fig:outputtaxo}
\end{figure*}

\subsubsection{Binary discrimination between known-topic and novel-topic documents}
\label{subsubsec:noveltyeval}
For each partial topic hierarchy, we provide an ablation analysis on the novelty detection performance of \proposed, to validate the effectiveness of the embedding techniques:
\textbf{local embedding} (LE) and \textbf{keyword-guided discriminative embedding} (DE).
Note that there do not exist term-level novelty labels, we instead use document-level novelty labels indicating whether a document belongs to one of the deleted ground-truth topic classes or not.
In other words, we indirectly evaluate the capability of novel topic detection based on the topic assignment of documents, obtained by Equation~\eqref{eq:docassign}.
We consider precision, recall, and F1 score as the evaluation metrics.
For a fair comparison, we select the optimal hyperparameter value $\beta\in\{1.5, 2.0, 2.5, 3.0\}$ to determine the novelty threshold (in Equation~\eqref{eq:novscore}) of the ablated methods.

In Table 5, \proposed that adopts both LE and DE performs the best for distinguishing novel-topic documents from known-topic counterparts.
Particularly, DE considerably improves the novelty detection performance, by collecting the given topic names from the sub-tree rooted at each node and utilizing them as the keywords for the topic, as discussed in Section~\ref{subsubsec:disemb}.
In summary, LE and DE optimize the embedding space further discriminative among known sub-topics, and it is helpful to enhance the binary discrimination between known and novel sub-topics as well.

\begin{figure}[t]
    \centering
    \includegraphics[width=\linewidth]{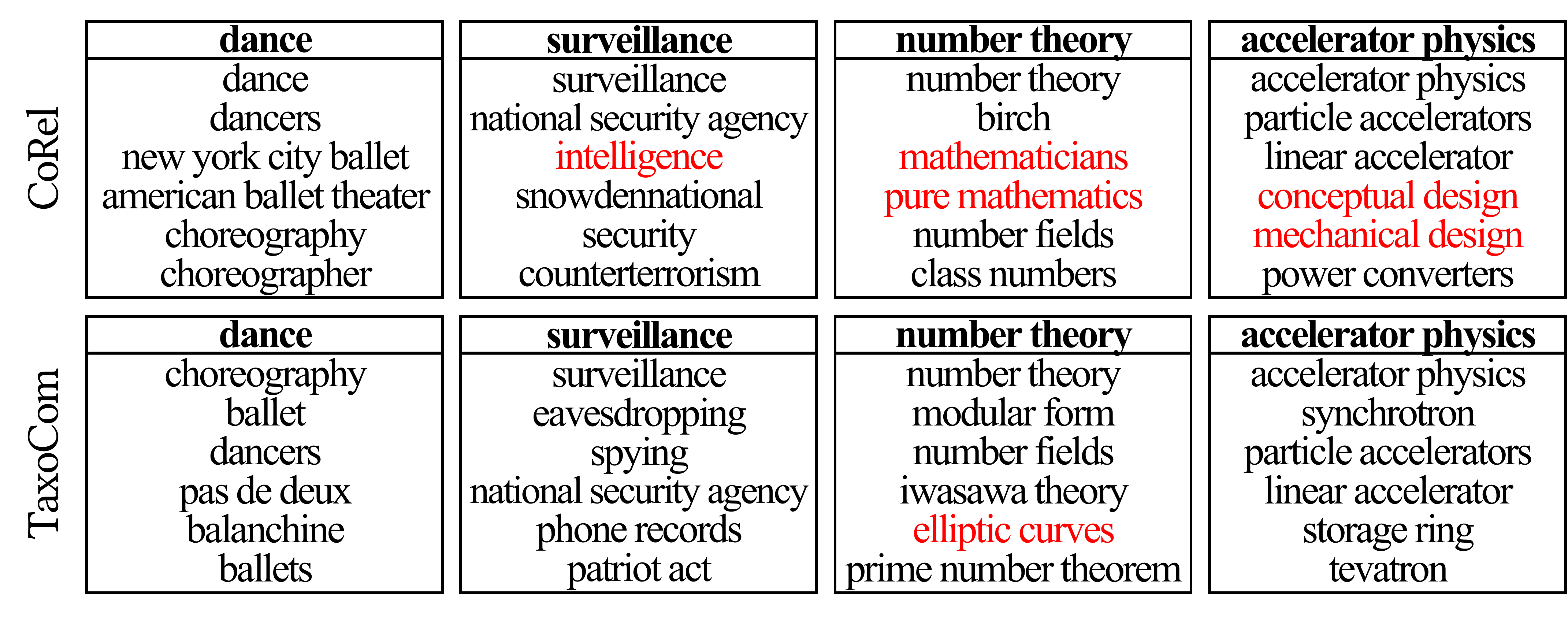}
    \caption{Topic terms retrieved by \corel and \proposed.}
    \label{fig:casestudy}
\end{figure}

\subsection{Qualitative Analysis}
\label{subsec:qualanal}
\subsubsection{Case study on topic taxonomy}
\label{subsubsec:casestudy}
We qualitatively examine the output taxonomy of \proposed.
Figure~\ref{fig:outputtaxo} shows that \proposed expands the topic taxonomy while preserving the high-level design of the given topic hierarchy $\taxoc$.
To be specific, in case of \nyt, it successfully identifies not only the first-level missing topic \textit{science} but also the second-level ones including \textit{music}, \textit{football}, and \textit{soccer}.
We observe that several center terms of novel topic nodes do not exactly match with the ground-truth topic names, such as \textit{spacecraft}-\textit{cosmos} (\nyt), \textit{data processing}-\textit{computer science} (\arxiv), and \textit{fluid mechanics}-\textit{fluid dynamics} (\arxiv).
Nevertheless, it is obvious that they represent the same conceptual topic of some documents in the text corpus, in light of the terms assigned to them.

Furthermore, we compare the topic nodes (and their anchor terms) identified by \proposed and \corel.
In Figure~\ref{fig:casestudy}, several topic terms of \corel are too general to belong to the topic (marked in red), whereas \proposed selectively filters the topic-relevant terms by taking advantage of topic sub-corpus.
In Table~\ref{tbl:casestudy}, \corel often finds redundant topics semantically overlapped with a known sub-topic (\ding{34}), or just novel entities in the ``is-a'' relation rather than a topic class of documents in the corpus ($\Motimes$).
Especially, \corel fails to find any of the first-level missing topics due to the lack of given ``is-a'' relations.
%Note that \josh is not able to identify novel topics at all.
In contrast, \proposed effectively completes the latent topic structure of the text corpus by discovering novel topics that semantically deviate from the known topics.

\begin{table}[t]
\caption{Novel sub-topics found by \corel and \proposed.}
\centering
\resizebox{0.99\linewidth}{!}{%
\begin{tabular}{ccUVV}
    \toprule
    \multicolumn{2}{c}{\multirow{2.5}{*}{{\shortstack{\textbf{Given}\\$\mathbf{\bm{\mathcal{H}^0}}$}}}} & \multirow{2}{*}{\textbf{Topic}} & \multicolumn{2}{c}{\textbf{Identified Novel Sub-topics}} \\\cmidrule(l){4-5}
    & & & \textbf{\corel} & \textbf{\proposed} \\\midrule
    \multirow{5}{*}{\rotatebox[origin=c]{90}{\nyt}}
    & $\taxoa$ & root & - & arts \\\cmidrule(l){2-5}
    & $\taxob$ & sports & baseball, softball($\Motimes$), squash($\Motimes$), wrestling($\Motimes$), ... & baseball, hockey \\\cmidrule(l){2-5}
    & $\taxoc$ & arts & jazz, hip hop, acting(\ding{34}), animation($\Motimes$), ballet(\ding{34}), ... & music \\\midrule
    \multirow{7}{*}{\rotatebox[origin=c]{90}{\arxiv}}
    & $\taxoa$ & root & - & natural sciences \\\cmidrule(l){2-5}
    & $\taxob$ & computer science & object detection, authentication(\ding{34}), database systems(\ding{34}), ... & code construction, game, object recognition, parallel~computing  \\\cmidrule(l){2-5}
    & $\taxoc$ & physics & astrophysics(\ding{34}), general~relativity(\ding{34}), particle physics(\ding{34}), ... & atomic theory, fluid~mechanics, electromagnetism(\ding{34})  \\
    \bottomrule
\end{tabular}
}
\label{tbl:casestudy}
\end{table}

\subsubsection{Visualization of discriminative embedding space}
\label{subsubsec:embvis}
We also visualize our spherical embedding space for the \nyt dataset, when $\taxoc$ is given as the partial topic hierarchy.
Figure~\ref{fig:embspace} shows the embedding space plotted by t-SNE~\cite{van2008visualizing} for (i) the root node and (ii) the \textit{sports} node.
In the left figures, the anchor terms assigned in multiple sub-topics are marked in different colors, while each center term and the non-anchor terms are plotted as black asterisks and white circles, respectively.
We annotate the center term of the novel clusters that correspond to the novel topic nodes (i.e., \textit{science}, \textit{football}, and \textit{soccer}) presented in Section~\ref{subsubsec:casestudy}.
On the other hand, the right figures illustrate the binary discrimination between known-topic and novel-topic terms, determined based on the novelty score (Equation~\eqref{eq:novscore}).
Our confidence-based novelty score is effective to detect novel-topic terms (and their dense clusters) clearly distinguishable from known-topic terms in the embedding space.
%In conclusion, for topic taxonomy completion, \proposed performs the hierarchical clustering that recursively tunes the local embedding space and finds out sub-topic clusters.

\section{Conclusion}
\label{sec:conc}
This paper studies a new problem, named topic taxonomy completion, which aims to complete the topic taxonomy initiated with a user-provided partial hierarchy.
The proposed \proposed framework performs the hierarchical discovery of novel sub-topic clusters;
it employs the text embedding and clustering tailored for effective discrimination between known sub-topics and novel sub-topics.
%presents a weakly supervised framework for topic taxonomy construction, named \proposed, which leverages a partial or incomplete hierarchy of topic surface names given as prior knowledge.
%To generate the complete topic taxonomy of an input text corpus, \proposed performs the hierarchical discovery of novel sub-topic clusters;
%it employs the text embedding and clustering tailored for effective discrimination between known sub-topics and novel sub-topics.
%For each topic node, terms are selectively assigned into one of its child nodes considering both their relevance and representativeness to each sub-topic.
The extensive experiments show that \proposed successfully outputs the high-quality topic taxonomy which accurately matches with the ground-truth topic hierarchy.

In future work, we would like to enhance the center term of novel topic nodes so that it can best summarize the high-level concepts of its sub-topic clusters.
To this end, it would be interesting to (i) incorporate relation extraction techniques and (ii) employ a pre-trained language model;
this could be a bridge technique between an entity taxonomy and a topic taxonomy, with a comprehensive understanding of entities, relations, documents, and topics.

\begin{acks}
This work was supported by the IITP grant (No. 2018-0-00584, 2019-0-01906), the NRF grant (No. 2020R1A2B5B03097210), the Technology Innovation Program (No. 20014926), and the Korea Health Technology R\&D Project (HI18C2383).
It was also supported by US DARPA KAIROS Program (No. FA8750-19-2-1004), SocialSim Program (No. W911NF-17-C-0099), INCAS Program (No. HR001121C0165), National Science Foundation (IIS-19-56151, IIS-17-41317, IIS 17-04532), and the Molecule Maker Lab Institute: An AI Research Institutes program (No. 2019897).
\end{acks}

\bibliographystyle{ACM-Reference-Format}
\bibliography{BIB/bibliography}

\newpage
\appendix
\section{Supplementary Material}
\subsection{Pseudo-code of \proposed}
\label{subsec:pseudocode}
Algorithm~\ref{alg:overview} describes the overall process of our framework for topic taxonomy completion.
Starting from the root node (Line 2), \proposed recursively performs text embedding and clustering for each topic node (Lines 5 and 6) to find out multiple sub-topic term clusters.
Based on the clustering result, it expands the current topic taxonomy by inserting novel sub-topic nodes (Line 8).
In the end, \proposed outputs the complete topic taxonomy (Line 10).
\begin{algorithm}
    \DontPrintSemicolon
    \SetKwProg{Fn}{Function}{:}{}
    \SetKwComment{Comment}{$\triangleright$\ }{} 
	%\Comment*[r]{write comment here}
	
	\KwIn{An input text corpus $\docuset$ with its term set $\termset$, and a partial topic hierarchy $\taxo^0$} 
	%\KwOut{The complete taxonomy $\taxo$ and the set of terms and documents assigned to each node $\{\topicdocs{c}, \topicterms{c}\}$}
	
	\vspace{5pt}
	$\taxo \leftarrow \taxo^0$\;
	$q \leftarrow queue([(\taxo^0.rootNode, \termset, \docuset)])$ \;
    \While{$not$ $q.isEmpty()$}{
    $(c, \topicterms{c}, \topicdocs{c}) \leftarrow q.pop()$ \Comment*[r]{Current node}
    %$\topicdocs{c}^* \leftarrow$ \leftarrow RetrieveRelevantDocs($\topicterms{c}, \topicdocs{c};\docuset$)\;
    $\mathcal{E}_{c} \leftarrow$ \textsc{{Embedding}}($\topicterms{c},\topicdocs{c};\taxo$)  \Comment*[r]{Section~\ref{subsec:embedding}}
    $\mathcal{R}_c\leftarrow$ \textsc{{Clustering}}($\mathcal{E}_{c},\topicterms{c},\topicdocs{c};\taxo$) \Comment*[r]{Section~\ref{subsec:clustering}}
    \For{$(s, \topicterms{s}, \topicdocs{s})\in \mathcal{R}_c$}{
            $\taxo.updateChildNodes(c, (s, \topicterms{s}, \topicdocs{s}))$\;
            $q.push((s, \topicterms{s}, \topicdocs{s}))$\;
        }
    }
    \Return $\taxo$
    
\caption{The overall process of \proposed.}
\label{alg:overview}
\end{algorithm}

% \begin{algorithm}
%     \DontPrintSemicolon
%     \SetKwProg{Fn}{Function}{:}{}
%     \SetKwComment{Comment}{$\triangleright$\ }{} 
	
% 	\KwIn{A target topic $C$, the set of terms and documents $\topicterms{C}$ and $\topicdocs{C}$, and embedding vectors of the terms $\mathcal{U}$ } 
% 	\KwOut{$\topicterms{1}, \ldots, \topicterms{\numnc}$}
% 	\Comment*[r]{1. Novel Term Identification}
%     NovelTermIdentification \; 
%     \For{$\numnc \in [0, \ldots, |Child|]$}{
%     \Comment*[r]{2. Spherical Clustering}
%     \Comment*[r]{3. Anchor Term Selection}
%     \Comment*[r]{4. Mixture of vMF estimation}
%     }
%     \Return \{abcd\}
% \caption{Novelty adaptive clustering.}
% \label{alg:clustering}
% \end{algorithm}

\subsection{Implementation Details}
\label{subsec:impdetail}
We implement the main recursive process (Section~\ref{subsec:overview}) and the novelty adaptive clustering (Section~\ref{subsec:clustering}) by using Python,
and the locally discriminative embedding (Section~\ref{subsec:embedding}) is written in C for efficient optimization based on multi-threaded parallel computation.
For the other baselines, we employ the official author codes of \hlda\footnote{https://github.com/joewandy/hlda}, \taxogen\footnote{https://github.com/franticnerd/taxogen}, \josh\footnote{https://github.com/yumeng5/JoSH}, and \corel\footnote{https://github.com/teapot123/CoRel}.

For \textit{novelty adaptive clustering} of \proposed, we set $\beta=1.5, 3.0$ (for the first-level and second-level topic nodes, respectively) in the novelty threshold $\tau_{nov}$ (Equation~\eqref{eq:novscore}), and the significance threshold $\tau_{sig}=0.3$ (Equation~\eqref{eq:termassign}),
without further tuning for each dataset or initial topic hierarchy.
For \textit{locally discriminative embedding} of \proposed, we set the margin $m=0.3$ (Equation~\eqref{eq:josdopt}), and the number of neighbor terms $M=100$ (Section~\ref{subsubsec:locemb}) to retrieve the sub-corpus which are used for tuning the embedding space, i.e., top-$100$ closest term of each center term.
%\footnote{We empirically found that all these hyperparameters hardly affects the final output.}

For all the embedding-based methods (i.e., \taxogen, \josh, \corel, and \proposed) that optimize the Euclidean space or spherical space, we fix the number of negative terms (for each positive term pair) to 2 during the optimization.
For \hlda, we set (i) the smoothing parameter over document-topic distributions $\alpha=0.1$, (ii) the concentration parameter in the Chinese Restaurant Process $\gamma=1.0$, and (iii) the smoothing parameter over topic-word distributions $\eta=1.0$.
For \taxogen, we follow the parameter setting provided by~\cite{zhang2018taxogen};
i.e., the number of child nodes is set to 5.

% \subsection{Baseline Methods}
% \label{subsec:impdetail}
% Table~\ref{tbl:basecomp} presents the comparison of embedding-based baseline methods.

% \begin{table}[thbp]
% \caption{Comparison of embedding-based baseline methods.}
%     \centering
%     \resizebox{0.99\linewidth}{!}{%
%     \begin{tabular}{l|ccPPPP}
%         \toprule
%         \multirow{2}{*}{\textbf{Methods}} & \multirow{2}{*}{\shortstack{\textbf{Given}\\\textbf{Taxo}}} & \multirow{2}{*}{\shortstack{\textbf{Novel}\\\textbf{Topic}}} & \multicolumn{2}{c}{\textbf{Embedding}}& \multicolumn{2}{c}{\textbf{Term Mining}} \\
%         & & & LE & DE & Rel & Rep \\\midrule
%          \taxogen~\cite{zhang2018taxogen} & & & \checkmark  &  & \checkmark & \checkmark \\
%          \josh~\cite{meng2019weakly} & \checkmark & & & \checkmark & \checkmark &  \\
%          \corel~\cite{huang2020corel} & \checkmark  & \checkmark & & & \checkmark & \\
%          \textbf{\proposed} & \checkmark & \checkmark & \checkmark & \checkmark & \checkmark & \checkmark \\\bottomrule
%     \end{tabular}
%     }
%     \label{tbl:basecomp}
% \end{table}

\subsection{Constructed Topic Taxonomy}
\label{subsec:outputtaxo}

\begin{figure}[h]
\centering
\begin{subfigure}{\linewidth}
    \centering
    \includegraphics[width=\linewidth]{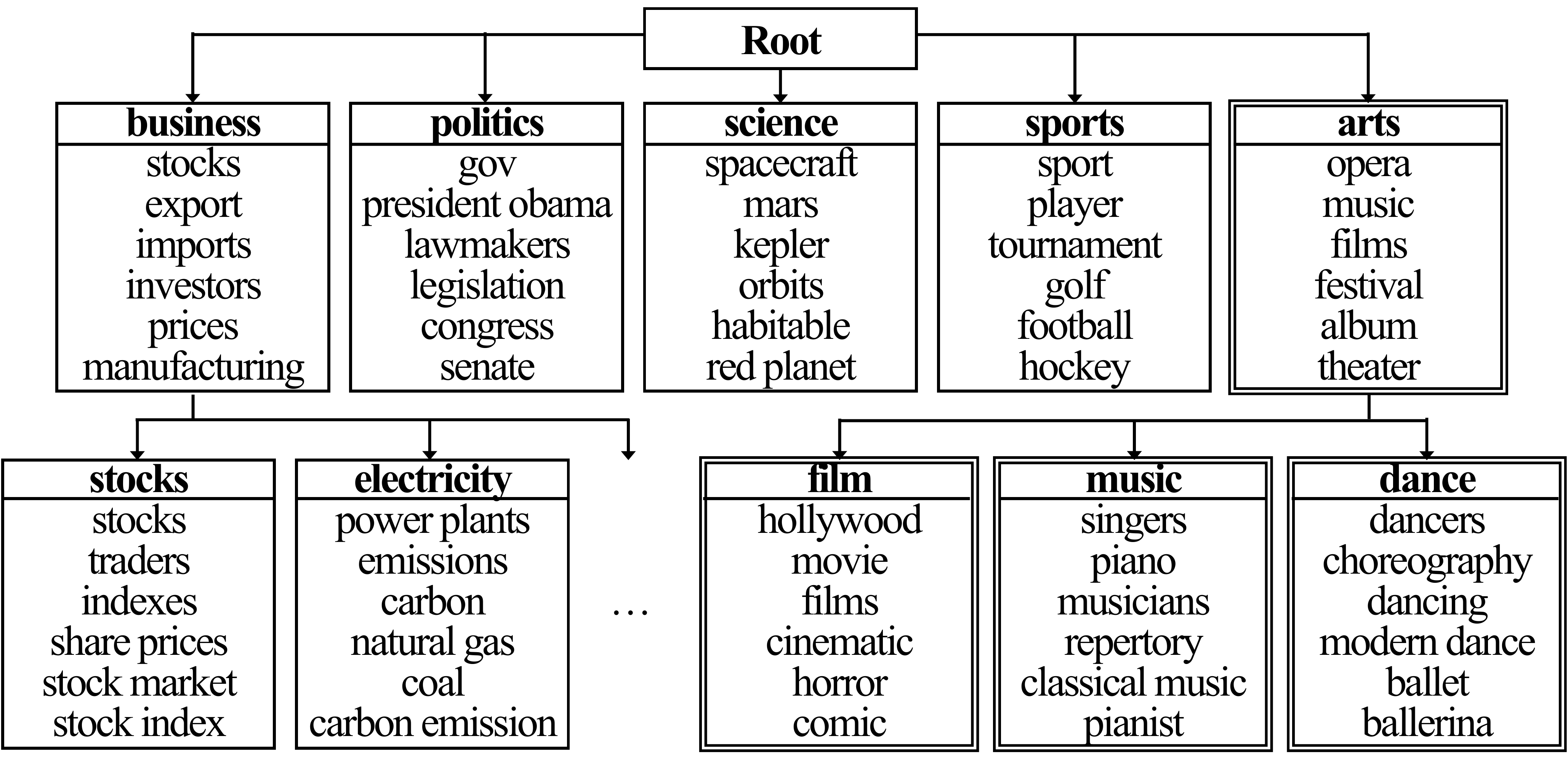}  
    \caption{Dataset: \nyt}
\end{subfigure}
\begin{subfigure}{\linewidth}
    \centering
    \includegraphics[width=\linewidth]{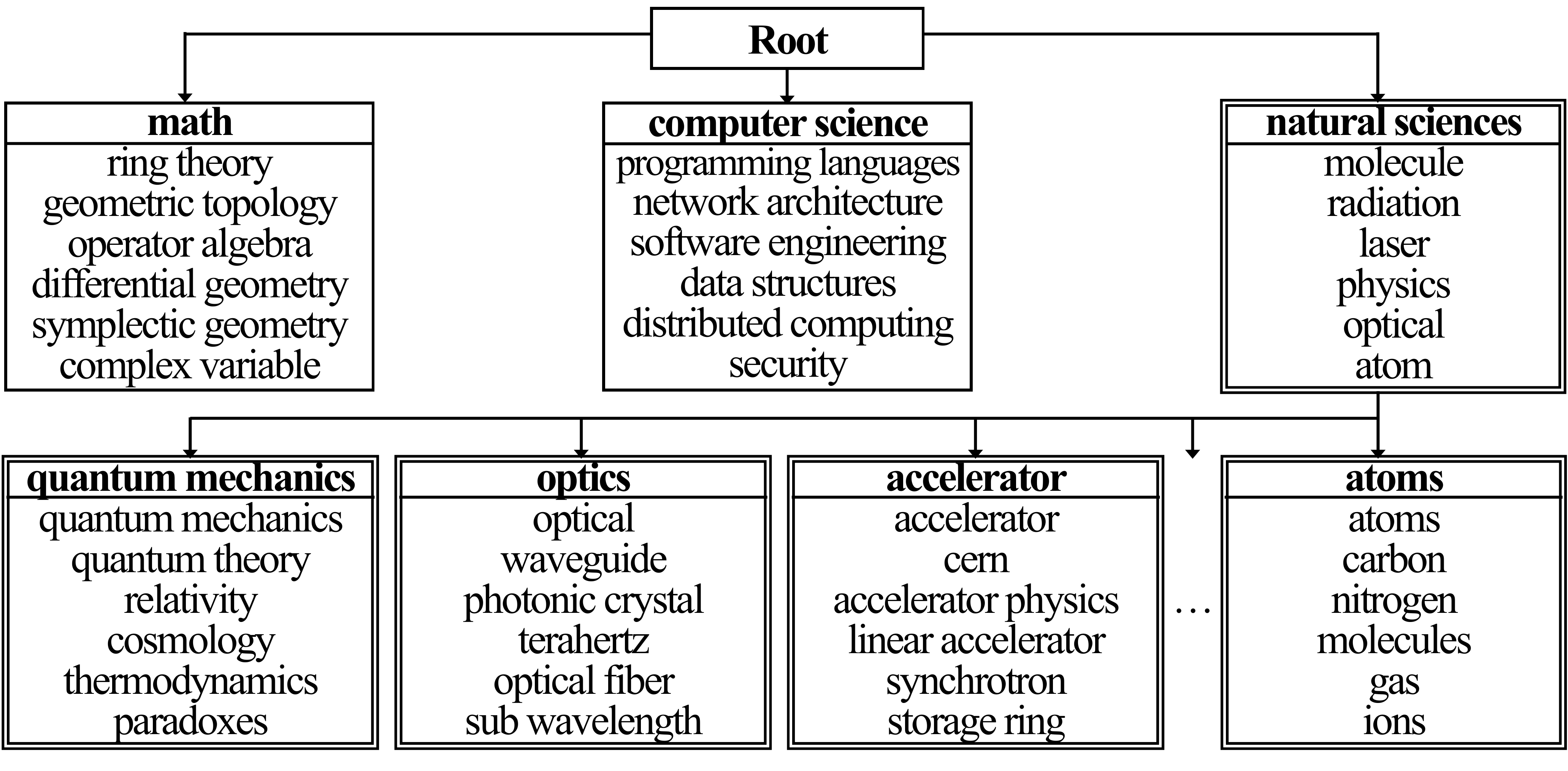}
    \caption{Dataset: \arxiv}
\end{subfigure}
\caption{The output taxonomy of \proposed, where $\taxoa$ is given as the initial topic hierarchy. Double-lined boxes represent the novel topic nodes.}
\label{fig:taxoa}
\end{figure}

\begin{figure}[t]
\centering
\begin{subfigure}{\linewidth}
    \centering
    \includegraphics[width=\linewidth]{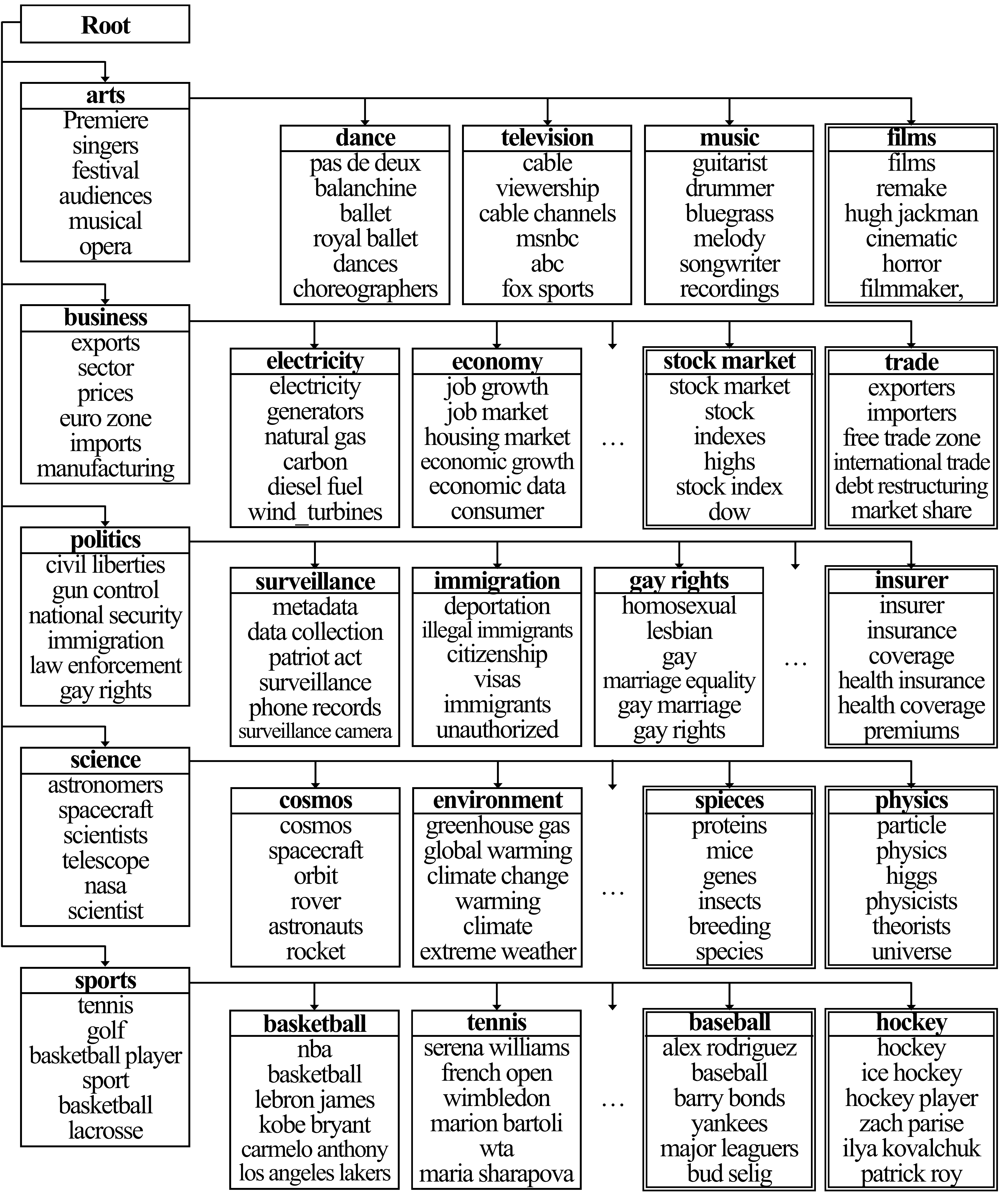}  
    \caption{Dataset: \nyt}
\end{subfigure}
\begin{subfigure}{\linewidth}
    \centering
    \includegraphics[width=\linewidth]{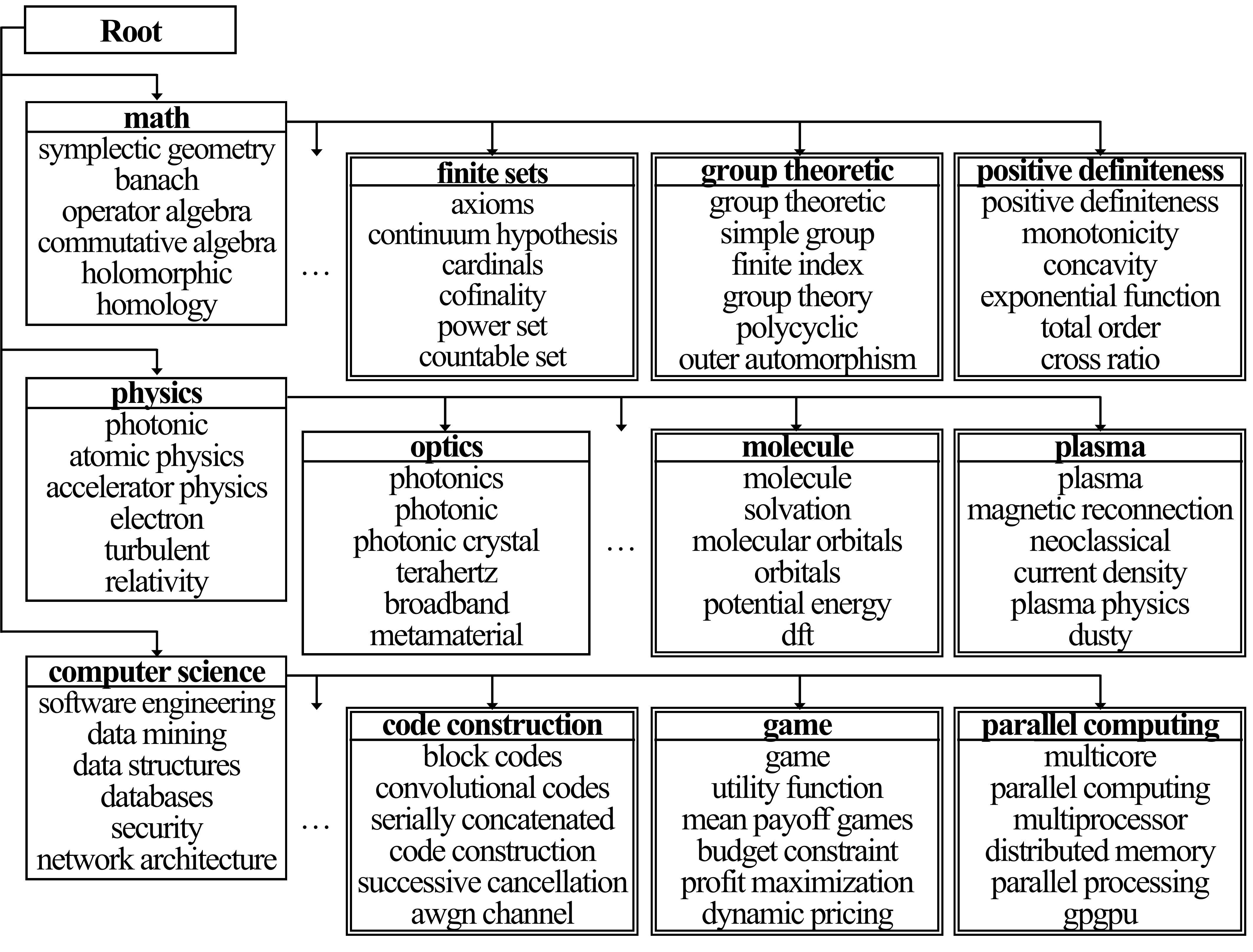}
    \caption{Dataset: \arxiv}
\end{subfigure}
\caption{The output taxonomy of \proposed, where $\taxob$ is given as the initial topic hierarchy. Double-lined boxes represent the novel topic nodes.}
\label{fig:taxob}
\end{figure}

Figures~\ref{fig:taxoa} and \ref{fig:taxob} show the topic taxonomies generated by \proposed, where $\taxoa$ and $\taxob$ are given as the initial topic hierarchy, respetively.\footnote{The output taxonomies started from $\taxoc$ is presented in Figure~\ref{fig:outputtaxo}.}
The center terms (i.e., topic names) and topic terms are presented without any post-processing such as manual filtering or selection that requires human labor.
Due to the space limit, the figures omit most of the second-level known topic nodes (i.e., the topics already included in $\taxoa$ or $\taxob$), rather focus on newly inserted novel topic nodes.
Note that the initial topic hierarchies are generated by random node drop of the ground-truth topic hierarchy:
a single first-level topic (in case of $\taxoa$) and a portion of second-level topics (in case of $\taxob$) is deleted, as listed in Table~\ref{tbl:partialtaxo}.

We can observe that \proposed effectively detects the deleted topic nodes and places them in the right position within the topic taxonomy.
In other words, the output taxonomies more completely cover the topic structure of each dataset, compared to the initial topic hierarchy, $\taxoa$ or $\taxob$.
The recursive clustering process of \proposed implicitly forces the hierarchical semantic relationship of parent-child node pairs, while clearly distinguishing novel sub-topic clusters from known sub-topic clusters based on the score of how confidently each term belongs to one of the known sub-topics (Equation~\ref{eq:novscore}).
As discussed in Section~\ref{subsubsec:casestudy}, some center terms of identified novel topic nodes do not match with the ground-truth topic names.
In spite of the minor mismatch, we can easily figure out the one-to-one mapping from the novel topics to the deleted ground-truth topics; 
for example, \textit{films}-\textit{movies}, \textit{stock market}-\textit{stocks}, \textit{insurer}-\textit{insurance} (\nyt), and \textit{natural sciences}-\textit{physics}, \textit{accelerator}-\textit{accelerator physics}, \textit{atoms}-\textit{atomic physics} (\arxiv).

% \subsection{Experimental Results}
% \label{subsec:addexp}
% \subsubsection{Qualitative comparison of top-10 terms}
% \label{subsubsec:qualcomp}

% \subsubsection{Parameter Analysis}
% \label{subsubbsec:paramanal}
% We investigate the performance changes of \proposed, with respect to the novelty threshold $\tau_{nov}$ and the significance threshold $\tau_{sig}$.
% (TO BE WRITTEN)

\end{document}